# Block Matching Frame based Material Reconstruction for Spectral CT

**Weiwen Wu**[1,2], **Qian Wang**[2], **Fenglin Liu**[1,3], **Yining Zhu**[4], **and Hengyong Yu**[2,*]

[1]Key Lab of Optoelectronic Technology and Systems, Ministry of Education, Chongqing University, Chongqing 400044, China

[2]Department of Electrical and Computer Engineering, University of Massachusetts Lowell, Lowell, MA 01854, USA

[3]Engineering Research Center of Industrial Computed Tomography Nondestructive Testing, Ministry of Education, Chongqing University, Chongqing 400044, China

[4]School of Mathematical Sciences, Capital Normal University, Beijing, 100048, China

The contributions of W. Wu and Q. Wang are equal.

*Corresponding author: Hengyong Yu

E-mail: hengyong-yu@ieee.org .

**Abstract:** Spectral computed tomography (CT) has a great potential in material identification and decomposition. To achieve high-quality material composition images and further suppress the x-ray beam hardening artifacts, we first propose a one-step material reconstruction model based on Taylor's first-order expansion. Then, we develop a basic material reconstruction method named material simultaneous algebraic reconstruction technique (MSART). Considering the local similarity of each material image, we incorporate a powerful block matching frame (BMF) into the material reconstruction (MR) model and generate a BMF based MR (BMFMR) method. Because the BMFMR model contains the L0-norm problem, we adopt a split-Bregman method for optimization. The numerical simulation and physical phantom experiment results validate the correctness of the material reconstruction algorithms and demonstrate that the BMF regularization outperforms the total variation and no-local mean regularizations.

## I. Introduction

The spectral computed tomography (CT) has attracted continuous attention for its outstanding performance in terms of tissue characterization, lesion detection and material decomposition (Kim *et al.*, 2015). The dual energy CT (DECT), as a simple version of spectral CT, has already been widely applied in many applications, such as material decomposition (Johnson *et al.*, 2007), abdomen angiography detection (Saito *et al.*, 2018; Liu *et al.*, 2016b), *etc*. The recent development of photon-counting detectors (PCDs) further enhances the prospect of spectral CT. This new-type of PCDs can distinguish the energy of each independently incoming x-ray photon by detecting the electronic pulse signal generated by the peak amplitude of quanta. Because each x-ray photon energy can be distinguished by thresholding the amplitude of quanta, the transmitted perturbative photon flux can be synchronously recorded within some small energy windows (Barber *et al.*, 2016). Theoretically, the measured data of different small channels can be utilized to reconstruct different attenuation maps for the same object (Alvarez and Macovski, 1976; Luo *et al.*, 2019; Luo *et al.*, 2017). Until now, the spectral CT scanners, which are equipped with PCDs whose energy windows are three or greater, have achieved great successes in contrast agent imaging and K-edge imaging (He *et al.*, 2012).

It is very common to use the energy information in the x-ray CT field (Hounsfield, 1973). For example, the DECT acquires two attenuation intensities using either two different x-ray spectra or two energy windows, and it reconstructs transmission intensity or material images using the collected datasets. Although only limited material component maps can be obtained from the DECT measurements (Johnson *et al.*, 2007), the technique is still of great significance because many materials to be reconstructed only contain two physical processes, i.e., photo-electric effect and Compton scattering. Regarding the material decomposition of DECT, it usually assumes two or three basis materials (Johnson *et al.*, 2007). However, it is difficult to satisfy the applications of vascular imaging of the head and neck (Symons *et al.*, 2018), dual-contrast agent (Muenzel *et al.*, 2016) and multiple contrast agent imaging (Symons *et al.*, 2017). These issues can be solved by utilizing the spectral CT by collecting projections from multiple energy bins (Symons *et al.*, 2018). Spectral CT can distinguish more basis materials by employing material decomposition methods (Zhang *et al.*, 2017). The decomposition methods can be divided into two categories: indirect and direct methods Those techniques are employed to detect cholesterol gallstones (Yang *et al.*, 2017), visual contrast media (Schirra *et al.*, 2014), improve tumor visibility (Brook *et al.*, 2012) and determine urinary stone composition (Li *et al.*, 2013).

The indirect material decomposition methods can further be divided into image-based and projection-based methods (Wang *et al.*, 2017). For the image-based methods, an intermediate step is performed to reconstruct the channel images from projections (Zhang *et al.*, 2017; Wu *et al.*, 2018b). Then, the material decomposition operates on the energy-channel images to obtain the final material maps. For the projection-based methods, the multi-energy projections are first decomposed to the sinograms for basis materials. Then, a conventional filtered backprojection or backprojection filtration method (Wu *et al.*, 2017b; Wu *et al.*, 2018a) is employed to reconstruct the material maps. As for the intermediate step of image-based methods, lots of iterative optimization techniques have been proposed (Taguchi and Iwanczyk, 2013), such as channel-by-channel reconstruction (Xu *et al.*, 2012) and joint spatial-spectral correlation reconstruction, tight frame sparsity (Zhao *et al.*, 2013), patch-based low-rank model (Kim *et al.*, 2015), HighlY constrained backPRojection  algorithm (Leng *et al.*, 2011), spectral prior image constraint compressed sensing  technique (Yu *et al.*, 2016), prior rank and sparsity model (Gao *et al.*, 2011), tensor dictionary learning  model (Zhang *et al.*, 2017) and its improved version (Wu *et al.*, 2018b), nonlocal low-rank and sparse matrix decomposition (Niu *et al.*, 2018), spatial-spectral cube matching frame (Wu *et al.*, 2018c), and non-local low-rank cube-based tensor factorization (Wu *et al.*, 2019).  Although the intensities of different energy channel images are different, they share the same structure of the scanned object. Thus, global low-rank, sparsity and tensor dictionary are efficient to characterize the structure similarity, leading to better reconstruction results. However, for the image domain material decomposition, there are two major unavoidable limitations: (1) because the weighted energy-averaged projections are used to reconstruct channel images, it fails to elaborate the real nonlinear relationship between the polychromatic projections and basic materials, resulting in reduced accuracy of the material decomposition (Brooks, 1977); (2) since the number of energy channels are often smaller than the material types for spectral CT, which means the reconstructed spectral channel images would be decomposed into a greater number of material categories,  it may have no unique solutions for material decomposition (Liu *et al.*, 2016a). The projection domain material decomposition first model the available projections by a high order decomposition function and then the image reconstruction techniques are followed. Actually, the combination of polychromatic projections should be consistent with the geometry (Flohr *et al.*, 2006). The biggest challenge in this case is data consistency among different energy bins. Any inconsistency, such as change in contrast agent concentration or internal pulsation can result in severe artifacts (Zhang *et al.*, 2008).

A direct material decomposition method, i.e., the one-step material reconstruction method (Barber *et al.*, 2016; Liu *et al.*, 2016a; Zhao *et al.*, 2015; Long and Fessler, 2014), can directly reconstruct material maps from the multi-energy datasets. For such one-step material reconstruction methods, the energy-dependent material intensity and the proportion of basis material or their product can be linearly modeled (Zhao *et al.*, 2015). There are two advantages. First, it can exactly describe the spectral imaging process to suppress x-ray beam hardening artifacts in the reconstructed images. Second, the optimization regularization penalty can be directly incorporated into the basis map reconstructions. However, the complexity of the energy spectral transmission model makes the decomposition process unstable, and the results are noticeably sensitive to noise. To improve the signal-to-noise ratio (SNR) of reconstructed material maps, the regularizations were incorporated into the one-step material decomposition model, such as total variation (TV) (Barber *et al.*, 2016), nonlocal TV (Liu *et al.*, 2016a), *etc*.

Because the image structure has self-similarity within a nonlocal region, a Block Matching 3D (BM3D) image model was proposed to characterize such self-similarity by grouping small similar image patches in a given search window (Dabov *et al.*, 2007). The BM3D algorithm was first introduced for image denoising. Due to its outstanding performance, it was later extended for image deblurring and inpainting (Danielyan *et al.*, 2012), *etc*. Recently, the BM3D frame was introduced to the MRI image reconstruction field as a powerful regularizer (Eksioglu, 2016).

In this paper, we focus on direct material reconstruction and propose a new one-step material simultaneous algebraic algebraic reconstruction technique (MSART) based on the first-order Taylor's expansion for spectral CT. Noting that the same material maps within nonlocal windows may share similar structures, we can improve the accuracy of material reconstruction by considering the structural similarities of material maps. Thus, to incorporate the similarity of material component maps and improve the robustness of the proposed material reconstruction model, the 3D block matching frame (Danielyan *et al.*, 2012) is employed as a regularizer for material reconstruction, leading to a Block Matching Frame Material Reconstruction (BMFMR) method. The contributions of this study are three-fold. First, to implement material reconstruction with the one-step procedure, we propose a direct material reconstruction model and develop the MSART based on Taylor's expansion. Second, to obtain better reconstruction results and improve the anti-noising ability of MSART, we incorporate the BMF regularizer into the MSART model and generate the BMFMR algorithm. Third, the split-Bregman method was employed to optimize the L0-norm optimization problem in the BMFMR model. The advantages of BMFMR method are mainly demonstrated in the following three aspects: i) directly obtaining the proportion of each basis material with a one-step procedure; ii) having a strong robustness and reducing image artifacts; iii) outperforming TV and non-local mean based material reconstruction methods in terms of image edge information preservation.

The rest of this paper is organized as follows. In section II, we present the model and solution of MSART for spectral CT. In section III, we briefly review the BM3D frame, establish the BMFMR model and adopt the split-Bregman to optimize our objective function. The algorithms and implementation details for comparison are also presented. In section IV, numerical simulations and physical phantom experiments are performed to evaluate the proposed algorithms. In section V, we discuss some related issues and make a conclusion.

## II. Material Map Reconstruction for Spectral CT

Let us consider a basic spectral imaging model. The photon number of an x-ray path $\ell$ from the $m^{th}$ energy window $E_m$ measured by a detector cell can be expressed as

$$w_{m\ell} = \int_{E_m} I_{m\ell}(E) e^{\int_{\mathbf{r}\in\ell} -\mu(E,\mathbf{r})d\mathbf{r}} dE \ . \tag{1}$$

Here $\int_{E_m} dE$ integrates over the range of $m^{th}$ energy channel and $\int_{\mathbf{r}\in\ell} d\mathbf{r}$ indicates integral over the x-ray path $\ell$. In this work, we assume the imaging object only contains $N$ materials, and $\mu(E,\mathbf{r})$ denotes the material linear attenuation coefficient for energy $E$ at position $\mathbf{r}$. $I_{m\ell}(E)$ represents the incident x-ray photon intensity emitting from the x-ray source for energy $E$.

To simplify Eq. (1), the x-ray attenuation coefficient $\mu(E,\mathbf{r})$ can be further expressed as (Schmidt *et al.*, 2017)

$$\mu(E,\mathbf{r}) = \sum_{n=1}^{N} \phi_n(E)\rho_n(\mathbf{r}), \tag{2}$$

where $\phi_n(E)$ is the basis function of the $n^{th}$ material at energy $E$ which can be determined by searching the tables in the National Institute of Standards and Technology (NIST) report (Hubbell and Seltzer, 1995), and $\rho_n(\mathbf{r})$ is the fraction of the $n^{th}$ material component at location $\mathbf{r}$. In this work, we will focus on reconstructing $\rho_n(\mathbf{r})$ which is referred to as material component maps. To avoid the inconsistency of energy bins, here we just employ the incident x-ray photon flux $I_{m\ell}^{(0)}$ at the $m^{th}$ energy bin rather than the total photon $I_\ell^{(0)}$ at all bins. Furthermore, $I_{m\ell}^{(0)}$ can be defined as

$$I_{m\ell}^{(0)} = \int_{E_m} I_{m\ell}(E) dE. \tag{3}$$

Substituting Eqs. (2) and (3) into Eq. (1), we have

$$S_{m\ell} = \int_{E_m} s_{m\ell}(E) e^{-\int_{\mathbf{r}\in\ell}\sum_{n=1}^{N}\phi_n(E)\rho_n(\mathbf{r})d\mathbf{r}} dE, \tag{4}$$

where $S_{m\ell} = w_{m\ell}/I_{m\ell}^{(0)}$, and $s_{m\ell}(E)$ represents the normalized energy spectrum distribution of x-ray intensity and detector sensitivity $s_{m\ell}(E) = \frac{I_{m\ell}(E)}{I_{m\ell}^{(0)}}$. Considering the orders of integral and summation in the exponential function in Eq. (4) are exchangeable, there will be

$$S_{m\ell} = \int_{E_m} s_{m\ell}(E) e^{-\sum_{n=1}^{N}\phi_n(E)\int_{\mathbf{r}\in\ell}\rho_n(\mathbf{r})d\mathbf{r}} dE. \tag{5}$$

Proceeding with a multi-energy CT model, we discretize Eq. (5) over the energy spectrum $E_m$

$$\begin{aligned} S_{m\ell} &\approx \sum_i s_{m\ell}(E_i) e^{-\sum_{n=1}^{N}\phi_n(E_i)\int_{\mathbf{r}\in\ell}\rho_n(\mathbf{r})d\mathbf{r}} \Delta E_i \\ &\approx \sum_i s_{m\ell}(E_i) e^{-\sum_{n=1}^{N}\phi_n(E_i)\mathbf{A}_{\ell\#}\boldsymbol{f}_n} \Delta E_i \\ &= \sum_i s_{m\ell}(E_i) e^{-\sum_{n=1}^{N}\phi_n(E_i)p_{n\ell}} \Delta E_i, \end{aligned} \tag{6}$$

where $p_{n\ell} = \mathbf{A}_{\ell\#}\boldsymbol{f}_n = \int_{\mathbf{r}\in\ell}\rho_n(\mathbf{r})d\mathbf{r}$, $\boldsymbol{f}_n$ is the vectorization of the reconstructed $n^{th}$ material map, $\mathbf{A}$ presents projection matrix in $\mathcal{R}^{L\times J}$ with $J=J_1\times J_2$, $J_1$ and $J_2$ represent width and height of the material maps, $L$ is the number of total x-ray paths, and $\mathbf{A}_{\ell\#}$ presents the $\ell^{th}$ row of $\mathbf{A}$. Then, we perform a logarithm operation on both sides of Eq. (6) and obtain

$$\overline{q_{m\ell}} = \ln S_{m\ell} \approx \ln\left(\sum_i s_{m\ell}(E_i) e^{-\sum_{n=1}^{N}\phi_n(E_i)p_{n\ell}} \Delta E_i\right), \tag{7}$$

Because the summation locates inside of the logarithm operation, it is difficult to solve this problem. Inspired by the E-ART reconstruction algorithm for dual-energy CT (Zhao *et al.*, 2015), we unfold the right hand of Eq. (7) with the first-order of Taylor's expansion at the current point $p_{n\ell}^{(k)}$

$$\overline{q_{m\ell}} = \ln\left(\sum_i s_{m\ell}(E_i)e^{-\sum_{n=1}^{N}\phi_n(E_i)p_{n\ell}^{(k)}}\Delta E_i\right) -$$

$$\sum_{n=1}^{N}\left(\frac{\left(\sum_i s_{m\ell}(E_i)\phi_n(E_i)e^{-\sum_{n=1}^{N}\phi_n(E_i)p_{n\ell}^{(k)}}\right)}{\left(\sum_i s_{m\ell}(E_i)e^{-\sum_{n=1}^{N}\phi_n(E_i)p_{n\ell}^{(k)}}\right)}\left(p_{n\ell} - p_{n\ell}^{(k)}\right) + \mathcal{O}\left(p_{n\ell} - p_{n\ell}^{(k)}\right)\right), \quad (8)$$

where $\mathcal{O}\left(p_{n\ell} - p_{n\ell}^{(k)}\right)$ presents Taylor's 2$^{\text{nd}}$ order infinitesimal and $k$ indicates the current iteration step. Thus, Eq. (8) can be simplified as

$$\overline{q_{m\ell}} = q_{m\ell}^{(k)} - \sum_{n=1}^{N}\left(\frac{\Theta_{m\ell n}^{(k)}}{s_{m\ell}^{(k)}}\left(p_{n\ell} - p_{n\ell}^{(k)}\right) + \mathcal{O}\left(p_{n\ell} - p_{n\ell}^{(k)}\right)\right) \quad (9a)$$

where

$$q_{m\ell}^{(k)} = \ln\left(\sum_i s_{m\ell}(E_i)e^{-\sum_{n=1}^{N}\phi_n(E_i)p_{n\ell}^{(k)}}\Delta E_i\right), \quad (9b)$$

$$\Theta_{m\ell n}^{(k)} = \sum_i \phi_n(E_i)s_{m\ell}(E_i)\Delta E_i e^{-\sum_{n=1}^{N}\phi_n(E_i)p_{n\ell}^{(k)}}, \quad (9c)$$

$$s_{m\ell}^{(k)} = \sum_i s_{m\ell}(E_i)e^{-\sum_{n=1}^{N}\phi_n(E_i)p_{n\ell}^{(k)}}. \quad (9d)$$

Now, considering all energy channels, the vector form of Eq. (9a) can be expressed as

$$\begin{bmatrix}\overline{q_{1\ell}} - q_{1\ell}^{(k)}\\ \vdots \\ \overline{q_{M\ell}} - q_{M\ell}^{(k)}\end{bmatrix} = -\begin{bmatrix}S_{1\ell}^{(k)} & \cdots & 0\\ \vdots & \ddots & \vdots\\ 0 & \cdots & S_{M\ell}^{(k)}\end{bmatrix}^{-1} \times \begin{bmatrix}\Theta_{1\ell 1}^{(k)} & \cdots & \Theta_{1\ell N}^{(k)}\\ \vdots & \ddots & \vdots\\ \Theta_{M\ell 1}^{(k)} & \cdots & \Theta_{M\ell N}^{(k)}\end{bmatrix}\begin{bmatrix}p_{1\ell} - p_{1\ell}^{(k)}\\ \vdots\\ p_{N\ell} - p_{N\ell}^{(k)}\end{bmatrix} + \begin{bmatrix}\sum_{n=1}^{N}\mathcal{O}\left(p_{n\ell} - p_{n\ell}^{(k)}\right)\\ \vdots\\ \sum_{n=1}^{N}\mathcal{O}\left(p_{n\ell} - p_{n\ell}^{(k)}\right)\end{bmatrix}$$

.(10)

Eq. (10) is equivalent to

$$\begin{bmatrix}S_{1\ell}^{(k)}\sum_{n=1}^{N}\mathcal{O}\left(p_{n\ell} - p_{n\ell}^{(k)}\right)\\ \vdots\\ S_{M\ell}^{(k)}\sum_{n=1}^{N}\mathcal{O}\left(p_{n\ell} - p_{n\ell}^{(k)}\right)\end{bmatrix} = \begin{bmatrix}S_{1\ell}^{(k)} & \cdots & 0\\ \vdots & \ddots & \vdots\\ 0 & \cdots & S_{M\ell}^{(k)}\end{bmatrix}\begin{bmatrix}\overline{q_{1\ell}} - q_{1\ell}^{(k)}\\ \vdots\\ \overline{q_{M\ell}} - q_{M\ell}^{(k)}\end{bmatrix} + \begin{bmatrix}\Theta_{1\ell 1}^{(k)} & \cdots & \Theta_{1\ell N}^{(k)}\\ \vdots & \ddots & \vdots\\ \Theta_{M\ell 1}^{(k)} & \cdots & \Theta_{M\ell N}^{(k)}\end{bmatrix}\begin{bmatrix}p_{1\ell} - p_{1\ell}^{(k)}\\ \vdots\\ p_{N\ell} - p_{N\ell}^{(k)}\end{bmatrix},$$

(11)

where $S_{m\ell}^{(k)}$ ($m = 1, \ldots, M$) is a constant. To provide a better estimation for $p_{n\ell}$ from the measurement $\overline{q_{m\ell}}$ based on Eq. (8), we need to minimize the error term which is equivalent to the right side of Eq. (11). That is,

$$\min_{\boldsymbol{P}}\left\{\sum_{\ell=1}^{L}\left\|\boldsymbol{\Theta}_{\#\ell\#}^{(k)}\left(\boldsymbol{P}_{\#\ell} - \boldsymbol{P}_{\#\ell}^{(k)}\right) + \boldsymbol{S}_{\ell}^{(k)}\left(\overline{\boldsymbol{Q}}_{\#\ell} - \boldsymbol{Q}_{\#\ell}^{(k)}\right)\right\|_F^2\right\}, \quad (12)$$

where $\boldsymbol{P}_{\#\ell}$ is $\ell^{\text{th}}$ column of $\boldsymbol{P} = \begin{bmatrix} p_{11} & \cdots & p_{1L} \\ \vdots & \ddots & \vdots \\ p_{N1} & \cdots & p_{NL} \end{bmatrix}$ and represents the measured projection data over the full energy spectrum for the $\ell^{\text{th}}$ x-ray path, $\boldsymbol{\Theta}_{\#\ell\#}^{(k)} = \begin{bmatrix} \Theta_{1\ell 1}^{(k)} & \cdots & \Theta_{1\ell N}^{(k)} \\ \vdots & \ddots & \vdots \\ \Theta_{M\ell 1}^{(k)} & \cdots & \Theta_{M\ell N}^{(k)} \end{bmatrix}$, $\boldsymbol{S}_{\ell}^{(k)} = \begin{bmatrix} S_{1\ell}^{(k)} & \cdots & 0 \\ \vdots & \ddots & \vdots \\ 0 & \cdots & S_{M\ell}^{(k)} \end{bmatrix}$, $\overline{\boldsymbol{Q}}_{\#\ell} = \begin{bmatrix} \overline{q_{1\ell}} \\ \vdots \\ \overline{q_{M\ell}} \end{bmatrix}$ and $\boldsymbol{Q}_{\#\ell}^{(k)} = \begin{bmatrix} q_{1\ell}^{(k)} \\ \vdots \\ q_{M\ell}^{(k)} \end{bmatrix}$.

Substituting $p_{n\ell} = \boldsymbol{A}_{\ell\#}\boldsymbol{f}_n$ into Eq. (12), we have

$$\min_{\boldsymbol{\mathcal{F}}} \left\{ \sum_{\ell=1}^{L} \left\| \boldsymbol{\Theta}_{\#\ell\#}^{(k)} \left( (\boldsymbol{A}_{\ell\#}\boldsymbol{\mathcal{F}})^T - (\boldsymbol{A}_{\ell\#}\boldsymbol{\mathcal{F}}^{(k)})^T \right) + \boldsymbol{S}_{\ell}^{(k)} \left( \overline{\boldsymbol{Q}}_{\#\ell} - \boldsymbol{Q}_{\#\ell}^{(k)} \right) \right\|_F^2 \right\}, \tag{13}$$

where $\boldsymbol{\mathcal{F}} = [\boldsymbol{f}_1, \boldsymbol{f}_2, \cdots, \boldsymbol{f}_N]$, and $\boldsymbol{\mathcal{F}}^{(k)}$ represents the material component maps at the $k^{\text{th}}$ iteration. Eq. (13) can be directly used to reconstruct material maps, and it can be considered as a material reconstruction model. To solve the optimization problem of Eq. (13), we employ the constraint $\boldsymbol{P} = (\boldsymbol{A}\boldsymbol{\mathcal{F}})^T$,

$$\min_{\boldsymbol{\mathcal{F}}} \left\{ \sum_{\ell=1}^{L} \left\| \boldsymbol{\Theta}_{\#\ell\#}^{(k)} \left( \boldsymbol{P}_{\#\ell} - \boldsymbol{P}_{\#\ell}^{(k)} \right) + \boldsymbol{S}_{\ell}^{(k)} \left( \overline{\boldsymbol{Q}}_{\#\ell} - \boldsymbol{Q}_{\#\ell}^{(k)} \right) \right\|_F^2 \right\}, \text{s.t.} \boldsymbol{P} = (\boldsymbol{A}\boldsymbol{\mathcal{F}})^T. \tag{14}$$

Eq. (14) is a constrained optimization problem, and it can be converted into an unconstrained one

$$\min_{\boldsymbol{\mathcal{F}}, \boldsymbol{P}} \left\{ \sum_{\ell=1}^{L} \left\| \boldsymbol{\Theta}_{\#\ell\#}^{(k)} \left( \boldsymbol{P}_{\#\ell} - \boldsymbol{P}_{\#\ell}^{(k)} \right) + \boldsymbol{S}_{\ell}^{(k)} \left( \overline{\boldsymbol{Q}}_{\#\ell} - \boldsymbol{Q}_{\#\ell}^{(k)} \right) \right\|_F^2 + \lambda \| (\boldsymbol{A}\boldsymbol{\mathcal{F}})^T - \boldsymbol{P} \|_F^2 \right\}, \tag{15}$$

where $\lambda > 0$ is a constant to balance the two terms. Since Eq. (15) only contains two optimized variables, which can equal to solving the following two sub-problems:

$$\begin{cases} \min_{\boldsymbol{P}} \left\{ \sum_{\ell=1}^{L} \left\| \boldsymbol{\Theta}_{\#\ell\#}^{(k)} \left( \boldsymbol{P}_{\#\ell} - \boldsymbol{P}_{\#\ell}^{(k)} \right) + \boldsymbol{S}_{\ell}^{(k)} \left( \overline{\boldsymbol{Q}}_{\#\ell} - \boldsymbol{Q}_{\#\ell}^{(k)} \right) \right\|_F^2 + \lambda \| ((\boldsymbol{A}\boldsymbol{\mathcal{F}})^T)^{(k)} - \boldsymbol{P} \|_F^2 \right\} & (16a) \\ \min_{\boldsymbol{\mathcal{F}}} \| (\boldsymbol{A}\boldsymbol{\mathcal{F}})^T - \boldsymbol{P}^{(k+1)} \|_F^2. & (16b) \end{cases}$$

The goal of Eq. (16a) is to decompose the measured projection dataset to all material components, which can be considered as a projection decomposition model. To solve Eq. (16a), let us denote

$$Y(\boldsymbol{P}) = \left\{ \sum_{\ell=1}^{L} \left\| \boldsymbol{\Theta}_{\#\ell\#}^{(k)} \left( \boldsymbol{P}_{\#\ell} - \boldsymbol{P}_{\#\ell}^{(k)} \right) + \boldsymbol{S}_{\ell}^{(k)} \left( \overline{\boldsymbol{Q}}_{\#\ell} - \boldsymbol{Q}_{\#\ell}^{(k)} \right) \right\|_F^2 \right\} + \lambda \| ((\boldsymbol{A}\boldsymbol{\mathcal{F}})^T)^{(k)} - \boldsymbol{P} \|_F^2, \tag{17}$$

According to the ***Theorem 1*** in the Appendix *A.1*, $Y(\boldsymbol{P})$ can be minimized in the following iterative format

$$\left( \boldsymbol{P}_{\#1}^{(k+1)}, \cdots, \boldsymbol{P}_{\#L}^{(k+1)} \right) = \begin{pmatrix} \boldsymbol{P}_{\#1}^{(k)} - \beta_1 \left( \boldsymbol{\Theta}_{\#1\#}^{(k)}{}^T \boldsymbol{\Theta}_{\#1\#}^{(k)} + \lambda \boldsymbol{\mathcal{I}} \right)^{-1} \boldsymbol{S}_1^{(k)} \left( \overline{\boldsymbol{Q}}_{\#1} - \boldsymbol{Q}_{\#1}^{(k)} \right) \\ \vdots \\ \boldsymbol{P}_{\#L}^{(k)} - \beta_1 \left( \boldsymbol{\Theta}_{\#L\#}^{(k)}{}^T \boldsymbol{\Theta}_{\#L\#}^{(k)} + \lambda \boldsymbol{\mathcal{I}} \right)^{-1} \boldsymbol{S}_L^{(k)} \left( \overline{\boldsymbol{Q}}_{\#L} - \boldsymbol{Q}_{\#L}^{(k)} \right) \end{pmatrix}^T, \tag{18}$$

where $\beta_1$ and $\lambda$ are constant parameters, and $\boldsymbol{\mathcal{I}}$ is an identity transform. Particularly, $\boldsymbol{P}_{\#\ell}^{(k+1)}$ can be updated as

$$\boldsymbol{P}_{\#\ell}^{(k+1)} = \boldsymbol{P}_{\#\ell}^{(k)} - \beta_1 \left( \boldsymbol{\Theta}_{\#\ell\#}^{(k)}{}^T \boldsymbol{\Theta}_{\#\ell\#}^{(k)} + \lambda \boldsymbol{\mathcal{I}} \right)^{-1} \boldsymbol{S}_{\ell}^{(k)} \left( \overline{\boldsymbol{Q}}_{\#L} - \boldsymbol{Q}_{\#L}^{(k)} \right), \tag{19}$$

Now, we consider Eq. (16b), which is equivalent to

$$\min_{\mathcal{F}} \left\| A\mathcal{F} - \left(P^{(k+1)}\right)^T \right\|_F^2. \tag{20}$$

Eq. (20) can be viewed as a material map reconstruction model. Note that Eq. (20) is a convex quadratic optimization, it can be solved by the steepest descent method

$$\mathcal{F}^{(k+1)} = \mathcal{F}^{(k)} - \beta_2 A^T \left(A\mathcal{F}^{(k)} - \left(P^{(k+1)}\right)^T\right), \tag{21}$$

where $\beta_2$ is a relaxation factor. If the projections are sufficient, we can also employ the classical analytic reconstruction methods to directly solve Eq. (20).

From Eqs. (19) and (21), we can observe that the proposed material reconstruction algorithm can be divided into projection data decomposition and map reconstruction procedures. Because this algorithm can directly reconstruct the material maps instead of decomposing the reconstructed spectral image, we name it material simultaneous algebraic reconstruction technique (MSART). The major steps of the proposed MSART can be summarized as the following **Algorithm I**.

---

**Algorithm I: MSART**

**Input**: $\bar{Q}$; Normalized energy spectrum; $\beta_1$; $\beta_2$; $\lambda$;
1: Initialization: $\mathcal{F} \leftarrow 0$, $Q^{(0)} \leftarrow 0$, $P \leftarrow 0$; $k = 1$;
2: **Repeat**
**Part I: Projection decomposition**
3: **for** $\ell := 1,2,\cdots, L$ **do**
4:   **for** $m := 1,2,\cdots, M$ **do**
5:     $\Theta_{m\ell n}^{(k)} = \sum_i \left(\phi_n(E_i) s_{m\ell}(E_i) e^{-\sum_{n=1}^N \phi_n(E_i) p_{n\ell}^{(k)}}\right)$;
6:     $s_{m\ell}^{(k)} = \sum_i \left(s_{m\ell}(E_i) e^{-\sum_{n=1}^N \phi_n(E_i) p_{n\ell}^{(k)}}\right)$;
7:     $q_{m\ell}^{(k)} = \ln\left(\sum_i s_{m\ell}(E_i) e^{-\sum_{n=1}^N \phi_n(E_i) p_{n\ell}^{(k)}} \Delta E_i\right)$;
8:   **end for**
9: $P_{\#\ell}^{(k+1)} = P_{\#\ell}^{(k)} - \beta_1 \left(\left(\Theta_{\#\ell\#}^{(k)}\right)^T \Theta_{\#\ell\#}^{(k)} + \lambda \mathcal{I}\right)^{-1} S_\ell^{(k)} \left(\bar{Q}_{\#\ell} - Q_{\#\ell}^{(k)}\right)$;
10: **end for**
**Part II: Material image reconstruction**
11: $\mathcal{F}^{(k+1)} = \mathcal{F}^{(k)} - \beta_2 A^T \left(A\mathcal{F}^{(k)} - \left(P^{(k+1)}\right)^T\right)$;
12: $k = k + 1$;
13: Enforcing the positive constraint over $\mathcal{F}^{(k+1)}$;
14: **Until** convergence
**Output**: Material component map $\mathcal{F}$

---

## III. Block Matching Frame Material Reconstruction (BMFMR) Method

### III.A Block Matching Filtering Frame

The BM3D was proposed in (Dabov *et al.*, 2007) as a nonlocal image denoising tool based on adaptive high-order groupwise models. For better understanding of the BM3D model, we first review the BM3D flowchart, which can be divided into three steps: i) *Grouping by Matching*. Similar small blocks in a given image window are collected to form a group. These Blocks in each group are stacked together to build 3D data arrays. ii) *Collaborative hard-threshold filtering*. The noise in 3D data arrays are suppressed by a hard-threshold filtering on the groupwise spectrum coefficients in a transform domain. The invertible transform is performed on spectral coefficients to estimate all the grouped blocks, and then the estimated blocks are re-arranged into the original image positions. iii) *Aggregation*. The final image is obtained by weighting all blockwise estimations. The theorem of BM3D frame based matrix representation was introduced in (Danielyan *et al.*, 2012). Assuming there is a vectorized image $\in \mathcal{R}^J$, the analysis equation for the BM3D image model can be formulated as

$$\mathcal{W} = \Phi(y). \tag{22}$$

Here, $\mathcal{W} \in \mathcal{R}^B$ is the 3D groupwise spectrum to store the 3D transform domain coefficients for each group extracted from the image. $\Phi \in \mathcal{R}^{B \times J}$ ($B \gg J$) represents a frame, which transforms the image $y$ into groupwise spectrum space and provides an explicit expression of the BM3D analysis operation. The inverse transform matrix, which converts the groupwise spectrum coefficients into the image space, can be considered as another transform frame. Moreover, it can be implemented by a frame $\Psi \in \mathcal{R}^{J \times B}$ and thus the original signal can be recovered by

$$y = \Psi(\mathcal{W}). \tag{23}$$

The product $\Phi \Psi$ should be $\mathcal{I}_{B \times B}$. Both $\Phi$ and $\Psi$ mean operating the particular 3D transform on each extracted group. The 3D transform form can be chosen such that it can be divided into a pair of 2D-intrablock and 1D-interblock sub-transforms. These separable transforms can not only efficiently exploit the data structure of the 3D patch-based groups, but also greatly reduce the computational complexity compared to a non-separable transform. For more details, refer to (Danielyan *et al.*, 2012).

*III.B Algorithm model*

Although the BM3D frame was originally proposed for image denoising and deblurring (Danielyan *et al.*, 2012), it has been widely applied for other applications. For example, the BM3D was introduced to MRI reconstruction and obtained satisfied results compared with other competing methods (Eksioglu, 2016). Considering the similarity of nonlocal structures from each material component map, we introduce the BM3D frame into the MSART model to enhance the anti-noising ability, leading to the BMFMR algorithm. Based on Eq. (13), the BMFMR can be modeled as

$$\min_{\mathcal{F}} \left\{ \sum_{\ell=1}^{L} \left\| \Theta_{\#\ell\#}^{(k)} \left( (A_{\ell\#}\mathcal{F})^T - ((A_{\ell\#}\mathcal{F})^T)^{(k)} \right) + S_{\ell}^{(k)} \left( \overline{Q}_{\#\ell} - Q_{\#\ell}^{(k)} \right) \right\|_F^2 + \sum_{n=1}^{N} \kappa_n \|\Phi_n(f_n)\|_p \right\}, \tag{24}$$

where $\|\cdot\|_p$ represents the p-norm ($0 \leq p \leq 1$), $\kappa$ represents the regularization vector and $\kappa_n$ represents the regularization parameter of $n^{th}$ material reconstruction. In this work, we only consider the case $p = 0$, and $\kappa_n$ is selected as a constant vector. The solution of Eq. (24) can be obtained by adopting a similar procedure of Eq. (13) except for the material map reconstruction step. That is, Eq. (24) can be converted into two sub-problems

$$\begin{cases} \min_{P} \left\{ \sum_{\ell=1}^{L} \left\| \Theta_{\#\ell\#}^{(k)} \left( P_{\#\ell} - P_{\#\ell}^{(k)} \right) + S_{\ell}^{(k)} \left( \overline{Q}_{\#\ell} - Q_{\#\ell}^{(k)} \right) \right\|_F^2 + \lambda \|((A\mathcal{F})^T)^{(k)} - P\|_F^2 \right\}, & (25a) \\ \min_{\mathcal{F}} \left\{ \left\| A\mathcal{F} - \left( P^{(k+1)} \right)^T \right\|_F^2 + \sum_{n=1}^{N} \tau_n \|\Phi_n(f_n)\|_0 \right\}, & (25b) \end{cases}$$

where $\tau_n = \kappa_n/\lambda$ and $\tau$ can be expressed as $\kappa/\lambda$. Eq. (25a) is exactly the same as Eq. (16a), which has been solved in the previous section. Eq. (25b) contains the L$_0$-norm of 3D transform domain coefficients and it is an NP hard problem. Here, we introduce an auxiliary matrix $g_n \in \mathcal{R}^{J_1 \times J_2}$, which is a cell of the 3$^{rd}$ tensor $\mathcal{G}$. Then, Eq. (25b) can be read as

$$\min_{f_n} \left\| Af_n - \left( P_{n\#}^{(k+1)} \right)^T \right\|_F^2 + \tau_n \|\Phi_n(g_n)\|_0, s.t. \, g_n = f_n. \tag{26}$$

Now, Eq. (26) is a linear optimization problem with equality constraints. This can be further converted into an unconstrained linear optimization problem:

$$\min_{f_n, g_n, t_n} \left( \left\| Af_n - \left( P_{n\#}^{(k+1)} \right)^T \right\|_F^2 + \gamma_n \|f_n - g_n - t_n\|_F^2 + \tau_n \|\Phi_n(g_n)\|_0 \right). \tag{27}$$

Here, $t_n \in \mathcal{R}^{J_1 \times J_2}$ represents a cell of the 3$^{rd}$ error-feedback tensor $\mathcal{T}$, and $\gamma_n$ is the coupling parameter of $n^{th}$ material component between the data fidelity and difference terms. Because Eq. (27) contains three variable matrixes, it can be further divided into three steps using an alternating iterative strategy:

$$\min_{f_n} \left\| A f_n - \left(P_{n\#}^{(k+1)}\right)^T \right\|_F^2 + \gamma_n \left\| f_n - g_n^{(k)} - t_n^{(k)} \right\|_F^2, \quad (28a)$$

$$\min_{g_n} \left\| f_n^{(k+1)} - g_n - t_n^{(k)} \right\|_F^2 + \tau_n \|\Phi_n(g_n)\|_0, \quad (28b)$$

$$\min_{t_n} \left\| f_n^{(k+1)} - g_n^{(k+1)} - t_n \right\|_F^2. \quad (28c)$$

Because Eqs. (28a) and (28c) are convex quadratic optimization problems, their solution can be given by Eqs. (29) and (30),

$$f_n^{(k+1)} = f_n^{(k)} - \beta_2 A^T \left(A f_n^{(k)} - \left(P_{n\#}^{(k+1)}\right)^T\right) - \gamma_n \left(f_n^{(k)} - g_n^{(k)} - t_n^{(k)}\right), \quad (29)$$

$$t_n^{(k+1)} = t_n^{(k)} - \left(f_n^{(k+1)} - g_n^{(k+1)}\right). \quad (30)$$

Eq. (28b) is to update $g_n$ with respect to fixed $f_n^{(k+1)}$ and $t_n^{(k)}$, which can be treated as a denoising problem with an attenuation operation over the frame $\Phi_n$. Because Eq. (28b) contains the $L_0$–norm of 3D transform domain coefficients, a hard thresholding method is employed. Thus, the update of $g_n$ can be given as

$$g_n^{(k+1)} = \Psi_n \left\lfloor \Phi_n \left(f_n^{(k+1)} - t_n^{(k)}\right) \right\rfloor_{\tau_n}, \quad (31)$$

where the operator $\lfloor \cdot \rfloor_{\tau_n}$ denotes the hard thresholding operation, which can be calculated as follow (Setzer, 2011)

$$\left\lfloor \Phi_n \left(f_n^{(k+1)} - t_n^{(k)}\right) \right\rfloor_{\tau_n} = \begin{cases} 0, & \Phi_n \left(f_n^{(k+1)} - t_n^{(k)}\right) < \sqrt{\tau_n} \\ \Phi_n \left(f_n^{(k+1)} - t_n^{(k)}\right), & \Phi_n \left(f_n^{(k+1)} - t_n^{(k)}\right) \geq \sqrt{\tau_n} \end{cases}. \quad (32)$$

In fact, Eq. (32) can be simply treated as the BM3D denoising algorithm formulated by the frame notations (Danielyan *et al.*, 2012). For the proposed BMFMR technique, we utilize the shrinkage method Eq. (31) as an approximate solution for (28b).

### III.C Algorithm comparison

To demonstrate the advantages of BMF regularization term in image reconstruction, the total variation (TV) and non-local mean (NLM) regularizations are introduced into the model Eq. (13). As a result, we formulate TV based and NLM based material reconstruction methods, named as TVMR and NLMMR. The mathematic model of TVMR can be written as

$$\min_{\mathcal{F}} \left\{ \sum_{\ell=1}^{L} \left\| \Theta_{\#\ell\#}^{(k)} \left((A_{\ell\#}\mathcal{F})^T - ((A_{\ell\#}\mathcal{F})^T)^{(k)}\right) + S_\ell^{(k)} \left(\overline{Q}_{\#\ell} - Q_{\#\ell}^{(k)}\right) \right\|_F^2 + \sum_{n=1}^{N} \xi_n \mathrm{TV}(f_n) \right\}, \quad (33)$$

where $\xi_n$ is an element of vector $\xi$ and it represents the regularization parameter of the $n^{th}$ material reconstruction. To solve Eq. (33), we adopt the same procedure of BMFMR except for the material map reconstruction step. For the material map reconstruction, a similar strategy reported in (Wu *et al.*, 2017a) is employed.

### III.D Algorithm implementation details

The BMFMR method can be divided into two main procedures: projection decomposition and image reconstruction. Because the projection decomposition steps in BMFMR, NLMMR and TVMR are the same as the MSART method, we only summarized the material image reconstruction steps. The workflow of the BMFMR algorithm is summarized as Algorithm II. First, the projection data from different energy channels are decomposed into different material projection datasets. Then, each material map can be obtained from the corresponding material projection. Because different material components have different levels of noise, parameters should be optimized and selected to reconstruct different materials. The steps 11-16 are to reconstruct the material maps one by one. Regarding the update of step 12, the projection matrix $A$ is too large to calculate directly. Thus, it is necessary to divide it into two substeps:

$$f_n^{\left(k+\frac{1}{2}\right)} = f_n^{(k)} - \beta_2 A^T \left(A f_n^{(k)} - \left(P_{n\#}^{(k+1)}\right)^T\right), \tag{34}$$

$$f_n^{(k+1)} = f_n^{\left(k+\frac{1}{2}\right)} - \gamma_n \left(f_n^{(k)} - \mathbf{g}_n^{(k)} - \mathbf{t}_n^{(k)}\right), \tag{35}$$

where $f_n^{\left(k+\frac{1}{2}\right)}$ stands for the intermediate result. First, we calculate $f_n^{(k)} - \mathbf{g}_n^{(k)} - \mathbf{t}_n^{(k)}$ and store it. Then, $f_n^{\left(k+\frac{1}{2}\right)}$ can be updated by using Eq. (34). Finally, $f_n^{(k+1)}$ can be calculated according to Eq. (35). The steps 13–14 are the BM3D hard thresholding denoising process for the current estimation. These hard thresholding steps can be implemented by using a highly available online optimized BM3D denoising toolbox. However, we only employ the 3D transform shrinkage step and avoid the wiener collaborative step in the BM3D denoising filtering to enhance the ability of recovering image features. To accurately generate the two frames $\Psi_n$ and $\Phi_n$ for the $n^{\text{th}}$ material reconstruction, the noise level $\sigma$ is estimated from the current map. Because different material images correspond to different noise levels, different thresholds $\tau_n$ can effectively improve the denoising ability for Algorithm II. In our all experiments, except the hard thresholding coefficient column vector $\tau$, other parameters are the same as the openly available BM3D denoising toolbox.

---

**Algorithm II: BMFMR**

**Input**: $\overline{Q}$; Normalized energy spectrum; $\beta_1$; $\beta_2$; $\lambda$; $\gamma$; $\tau$;
1: Initialization: $\mathcal{F} \leftarrow 0$, $Q^{(0)} \leftarrow 0$, $P \leftarrow 0$; $k = 0$; $\mathcal{T} \leftarrow 0$; $\mathcal{G} \leftarrow 0$;
2: **Repeat**
**Part I: Projection decomposition**
**Part II: Material image reconstruction**
11: **for** $n := 1,2,\cdots,N$ **do**
12: $f_n^{(k+1)} = f_n^k - \beta_2 A^T \left(A f_n^{(k)} - \left(P_{n\#}^{(k+1)}\right)^T\right) - \gamma_n \left(f_n^{(k)} - \mathbf{g}_n^{(k)} - \mathbf{t}_n^{(k)}\right)$;
13: Generating two frames $\Psi_n$ and $\Phi_n$ using $f_n^{(k+1)} - \mathbf{t}_n^{(k)}$;
14: $\mathbf{g}_n^{(k+1)} = \Psi \lfloor \Phi(f_n^{(k+1)} - \mathbf{t}_n^{(k)}) \rfloor_{\tau_n}$;
15: $\mathbf{t}_n^{(k+1)} = \mathbf{t}_n^{(k)} - \left(f_n^{(k)} - \mathbf{g}_n^{(k+1)}\right)$;
16: **end for**
17: Enforcing the positive constraint over $\mathcal{F}^{(k+1)}$;
18: $k = k + 1$;
19: **Until** convergence
**Output:** Material component map $\mathcal{F}$

## IV. Experiments and Results

In this study, we first perform extensive numerical simulations to validate the correctness of our theory and the proposed algorithms. Then, physical phantom experiments are carried out to demonstrate their applications. The advantages of BMFMR method in terms of material map reconstruction are shown by comparing with the MSART, TVMR and NLMMR methods. To quantitatively evaluate the performance of all material reconstruction methods, the root means square error (RMSE), peak-signal-to-noise ratio (PSNR) and structural similarity (SSIM) are calculated. The RMSE, PSNR and SSIM of $n^{th}$ material can be defined as

$$\text{RMSE} = \sqrt{\frac{\sum_{j_2=1}^{J_2}\sum_{j_1=1}^{J_1}(\boldsymbol{f}_n - \boldsymbol{f}_n^*)^2}{J_1 \times J_2}}, \qquad (36)$$

$$\text{PSNR} = 20\log_{10}\left(\frac{\max(\boldsymbol{f}_n)}{\text{RMSE}}\right), \qquad (37)$$

$$\text{SSIM} = \frac{2c_{\boldsymbol{f}_n}c_{\boldsymbol{f}_n^*}(2\sigma_{\boldsymbol{f}_n\boldsymbol{f}_n^*} + e_2)}{\left(c_{\boldsymbol{f}_n}^2 + c_{\boldsymbol{f}_n^*}^2 + e_1\right)\left(\sigma_{\boldsymbol{f}_n}^2 + \sigma_{\boldsymbol{f}_n^*}^2 + e_2\right)}, \qquad (38)$$

where $\boldsymbol{f}_n^*$ represents the reference material image, $c_{\boldsymbol{f}_n}$ and $c_{\boldsymbol{f}_n^*}$ are the mean values of $\boldsymbol{f}_n$ and $\boldsymbol{f}_n^*$, $\sigma_{\boldsymbol{f}_n\boldsymbol{f}_n^*}$ represents the covariance between $\boldsymbol{f}_n$ and $\boldsymbol{f}_n^*$, $\sigma_{\boldsymbol{f}_n}$ and $\sigma_{\boldsymbol{f}_n^*}$ are the standard deviations of $\boldsymbol{f}_n$ and $\boldsymbol{f}_n^*$, $e_1$ and $e_2$ are constants. From Eq. (36), it can be inferred that the smaller the RMSE value is, the higher quality the reconstructed image is. For the index of PSNR, a high PSNR value means high quality of reconstruction results. In terms of SSIM, the value range is from 0 to 1. The closer to 1.0 the SSIM is, the better the image quality is. The $\max(\boldsymbol{f}_n)$ represents the maximum of $f_n$.

*A. Numerical Simulations*

*a). Experiment preparation*

A realistic mouse thorax phantom, containing three basis materials: bone, soft tissue and iodine (**Fig. 1**), is utilized in this study. As shown in **Fig. 1**, 1.2% iodine is introduced as a contrast agent. From **Fig. 1**, it can be seen that iodine contrast agent has relative complicated image structure. It is good to validate the performances of all methods in image edge preservation. Besides, the phantom has two lung regions in soft tissue component and they can be treated as mixture of soft tissue and air. These lung regions have abundant image features and details. It can benefit to evaluate the performances of material reconstruction methods. A polychromatic 50 kVp x-ray source is employed. **Fig. 2(a)** illustrates the corresponding normalized x-ray spectrum for this study. 341 uniform samples (from 16 keV to 50 keV) are extracted and divided into eight different energy channels: [16, 22) keV, [22, 25) keV, [25, 28) keV, [28, 31) keV, [31, 34) keV, [34, 37) keV, [37, 41) keV, [41, 50) keV. To evaluate the proposed algorithms, the attenuation curves of basic materials are given in **Fig. 2(b)**. From Fig. 2(b), we can observe that there is only one k-edge among the material attenuation curves around 33.7 keV for iodine. As for the scan geometry, the distances from source to PCD and rotation center are 180mm and 132mm, respectively. The PCD contains 512 detector cells, each of which covers a length of 0.1mm. 640 projections are uniformly collected over a full scan. In this study, the PCD is assumed ideal without considering the detector absorption efficiency, detector response, x-ray scatter and so on. To simulate Poisson noise, the photon numbers of each x-ray path are set as $10^5$, which is usually employed to simulate noise level for real applications (Schmidt *et al.*, 2017). The higher the photon number is, the lower the noise level is. The noisy projection datasets in Radon space are obtained by a post-logarithmic operation on the received photon numbers. All the reconstructed material component images are 512×512 matrixes, and each pixel covers an area of 0.075×0.075 mm$^2$. All the iterative algorithms are stopped after 40 iterations.

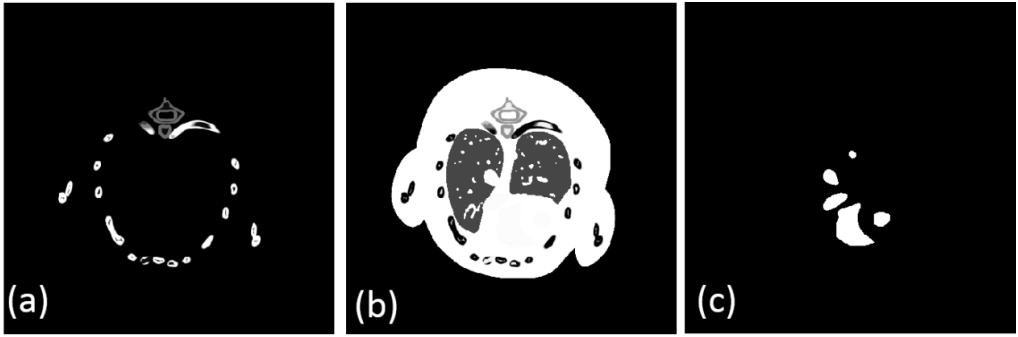

Figure 1. Three basic material maps of the realistic mouse thorax phantom. (a), (b) and (c) are bone, water and iodine material maps, respectively. The display windows for (a), (b) and (c) are [0 1], [0 1] and [0 0.012], respectively.

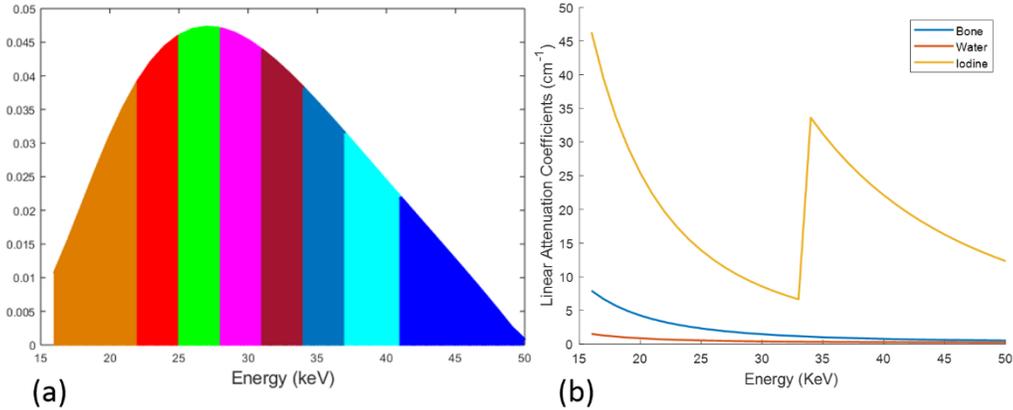

Figure 2. Normalized spectrum of a 50 kVp x-ray source used for CT simulation (left) and material-attenuation curves of three different basis materials (right).

Table I. Material image reconstruction parameters.

| | Bone | | | Water | | | Iodine | | |
|---|---|---|---|---|---|---|---|---|---|
| Numerical Simulation | $\tau_1$ | $\sigma_1$ | $\gamma_1$ | $\tau_2$ | $\sigma_2$ | $\gamma_2$ | $\tau_3$ | $\sigma_3$ | $\gamma_3$ |
| | 0.1 | 35 | 0.03 | 0.55 | 36 | 0.01 | 0.05 | 3 | 0.001 |
| | Aluminum | | | Water | | | Iodine | | |
| Physical Experiments | $\tau_1$ | $\sigma_1$ | $\gamma_1$ | $\tau_2$ | $\sigma_2$ | $\gamma_2$ | $\tau_3$ | $\sigma_3$ | $\gamma_3$ |
| | 0.8 | 35 | 0.2 | 0.6 | 25 | 0.2 | 0.004 | 3 | 0.02 |

The parameter selection is a challenging problem for iterative image reconstruction algorithms. In this study, both $\beta_1$ and $\beta_2$ are relaxation factors and they are set as 0.2. The Lagrangian multiplier $\lambda$ is fixed as 0.002 in all algorithms. To make it clear for the parameter selection of all the algorithms, other parameters are summarized in **Table I**.

*b). Material reconstruction*

Fig. 3 shows three basis material reconstructions. From **Fig. 3**, on one hand, we can observe that the proposed algorithms can reconstruct material maps directly from projection datasets, validating the correctness of our theoretical results. On the other hand, the BMFMR method obtains the best image quality compared with the MSART, TVMR and NLMMR algorithms. Because we decompose the projections into different material components, the noise levels can be easily magnified, resulting in magnified artifacts in the reconstructed material maps. Fortunately, these artifacts can be effectively reduced by certain prior regularization information, such as TV, NLM or BMF. In terms of bone component, the result of MSART contains some noise, resulting in degraded image quality. Compared with the MSART method, the TVMR and NLMMR methods can improve image quality to some extent by suppressing noise, and this can benefit to image edge preservation and features recovery. However, it can be observed that image features and edges are blurred in the TVMR and NLMMR results in **Fig. 3**, which makes it difficult to discriminate specific image structure in some cases. Compared with the TVMR and NLMMR methods, BMFMR can provide higher quality of bone maps with much clearer image edges. This point can be confirmed by the image structure indicated by red arrow "1" of the magnified ROI "A" in the first column of **Fig. 4**. In terms of soft tissue, the image edges and features from the MSART are submerged by severe noise. Compared with the MSART method, the TVMR, NLMMR and BMFMR methods can improve image quality by incorporating prior information into the material reconstruction model. Specifically, the features and image edges can be recovered by the TVMR method to some extent. However, the TVMR results still contain severe noises, resulting in artifacts that may be treated as image structures in some cases. Besides, the edges and features of soft tissue map are affected by blocky artifacts. It is hard to discriminate clear edges, which can be confirmed by image structures indicated by the red arrow "2" and "3" in **Fig. 4**. In contrast, the NLMMR can further reduce noise without blocky artifacts. However, the image features and details are missing in the soft tissue results. Compared with the TVMR and NLMMR results, BMFMR can provide more image features with clear image edges, which play an important role in distinguishing anatomical structures. These advantages can be further highlighted by image structures in the magnified ROIs "B", "C" and "D". In terms of iodine contrast agent, the noise is still distributed in a uniform region of the MSART results. The TVMR can reduce noise but with blurred image edges. Regarding the NLMMR result, it can be seen that it has a clear image edge. However, it still contain noise. Compared with these three comparisons, our proposed BMFMR method can provide clear image edge with reduced noise. Regarding the color rendering images, the magnified regions indicated by "E" in **Fig. 4** further confirm the outstanding performance of the BMFMR method in terms of image edge preservation.

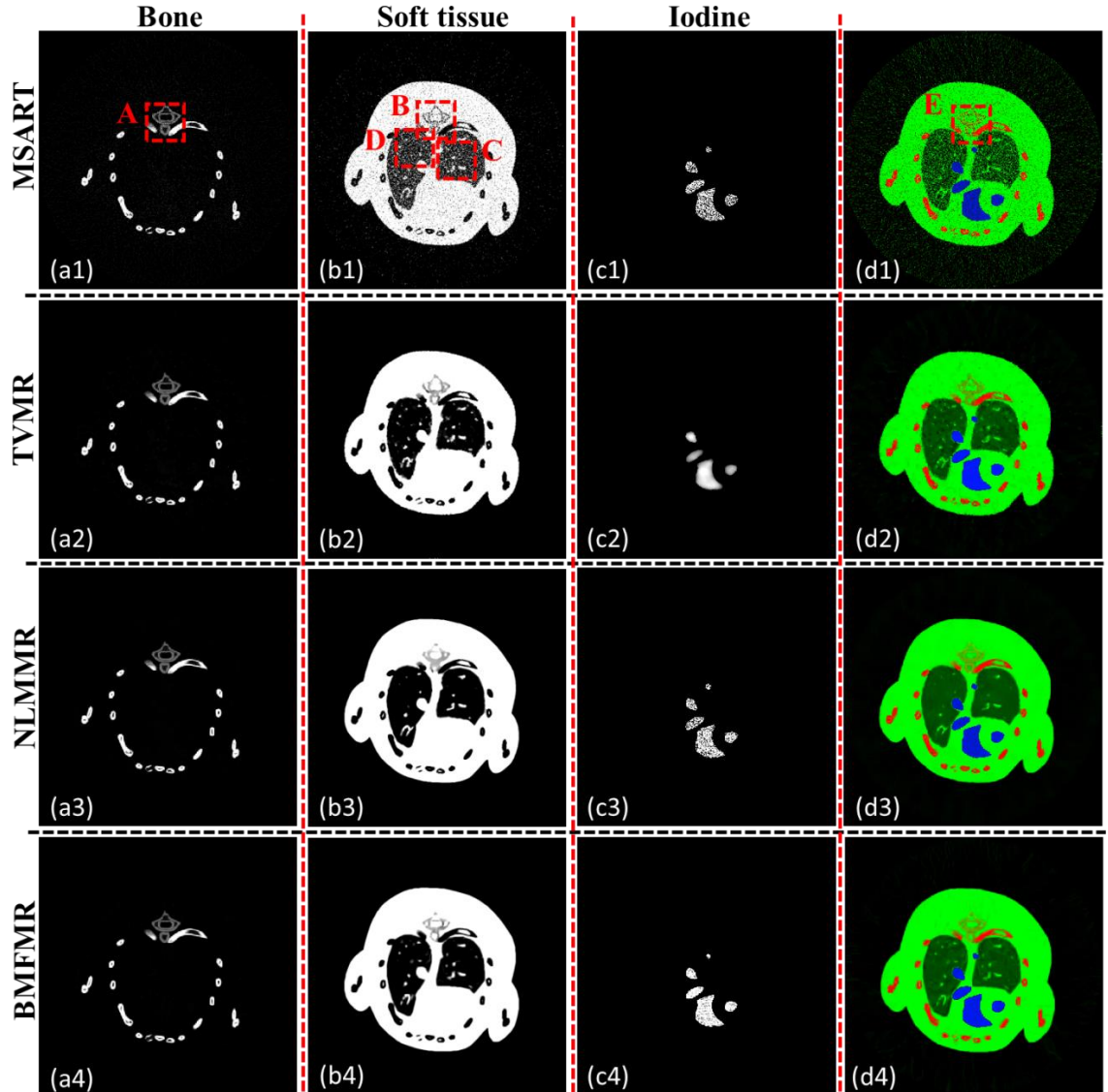

Figure 3. Reconstructed basis material images from 640 projections with $10^5$ photons for each x-ray path. The display windows of bone, water and iodine are [0.012 0.1], [0.35 0.80] and [0.011 0.012], respectively. The 4$^{th}$ column is the color rendering image, where blue, green and red represent the iodine contrast agent, soft tissue and bone, respectively.

To further demonstrate the advantages of the BMFMR method in material decomposition, **Fig. 5** shows the reconstructed difference images with respect to the ground truths. From **Fig. 5**, one can see the MSART reconstructions are strongly affected by noise and the difference images have largest errors. The TVMR, NLMMR and BMFMR methods can reduce the artifacts with smaller errors. Regarding the reconstructed bony image results, the residual of TVMR result is larger than that obtained by the NLMMR and BMFMR methods. In terms of soft tissue maps, it can be observed that the TVMR results contain many blotchiness artifacts, which are caused by noise distribution of TVMR results. Those artifacts may be treated as small image features in some cases. However, high accuracy can be obtained for soft tissue images by reducing these blotchiness artifacts to some extent in the NLMMR and BMFMR results. In fact, more image details and features are missing in NLMMR results. Regarding the iodine contrast agent, since the TV regularization focuses on penalizing the amplitude of gradient image, the uniformity of TVMR results seems better than that obtained by the NLMMR and BMFMR methods. However, the profiles in the TVMR results are also blurred. Correspondingly, the NLMMR can provide clearer image edges but also contains salient regions. The BMFMR can improve image quality.

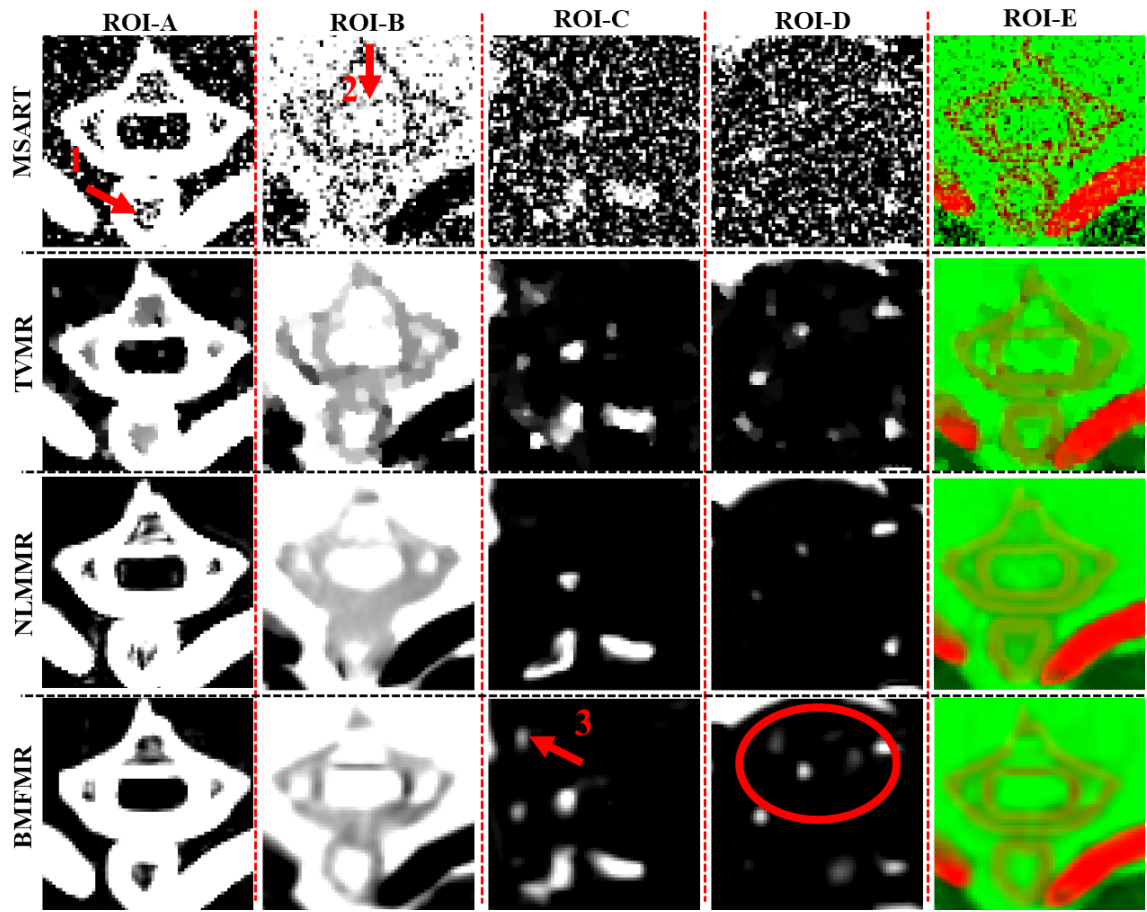

Figure 4. The magnified ROIs A, B, C, D and E in Fig. 3.

To quantitatively evaluate the accuracy of material reconstruction of all algorithms, the RMSE, SSIM and PSNR are computed and listed in **Table II**. Because the quantum noises are superimposed into the projection domain pixel by pixel, those pixel-wise noises are independent. That is, we have imposed almost half million pixel-wise independent noises into the projections. With the linear operation of backprojection, the pixel-wise noises in the reconstructed image can also be viewed as independent. When the RMSE, SSIM and PSNR values are calculated over a reconstructed image, they can be viewed as the statistical values over the whole image for the pixel-wise independent noise. Due to the structure bone component is simple, NLMMR and BMFMR can obtain similar higher accuracy of material component than that obtained using MSART and TVMR methods in terms of RMSE and PSNR. Furthermore, the measurement of SSIM index, which usually measures the similarity between two images, demonstrates the bony component from BMFMR method is closer to ground truth than those obtained by other competitors. As for soft tissue component, the MSART method has poor performance and thus it has the largest RMSE value and smallest SSIM and PSNR values. Compared with MSART, TVMR and NLMMR can improve the material image quality with lower RMSE values and higher PSNR and SSIM values. However, the image features are missing in NLMMR results. TVMR can recover material images with more image features than NLMMR even BMFMR in some cases, however the image edge of TVMR method usually contains noise and artifacts so that it is difficult to discriminate image features or noise. Fortunately, our proposed BMFMR can make a good tradeoff between image quality and quantitatively analysis. From **Table II**, it can be seen that the BMFMR method has the smallest RMSE and highest SSIM and PSNR values. For iodine contrast agent component, BMFMR can obtain the smallest RMSE values and highest PSNR values than other comparisons. As we have aforementioned that the TVMR can provide uniform component distribution than those achieved by NLMMR and BMFMR methods, it result in the SSIM value from TVMR is slightly higher than NLMMR and BMFMR. We also need to note that the image edges of iodine contrast agent are blurred in iodine contrast iodine agent result. All of these quantitative results further confirm the advantages of our proposed BMFMR algorithm.

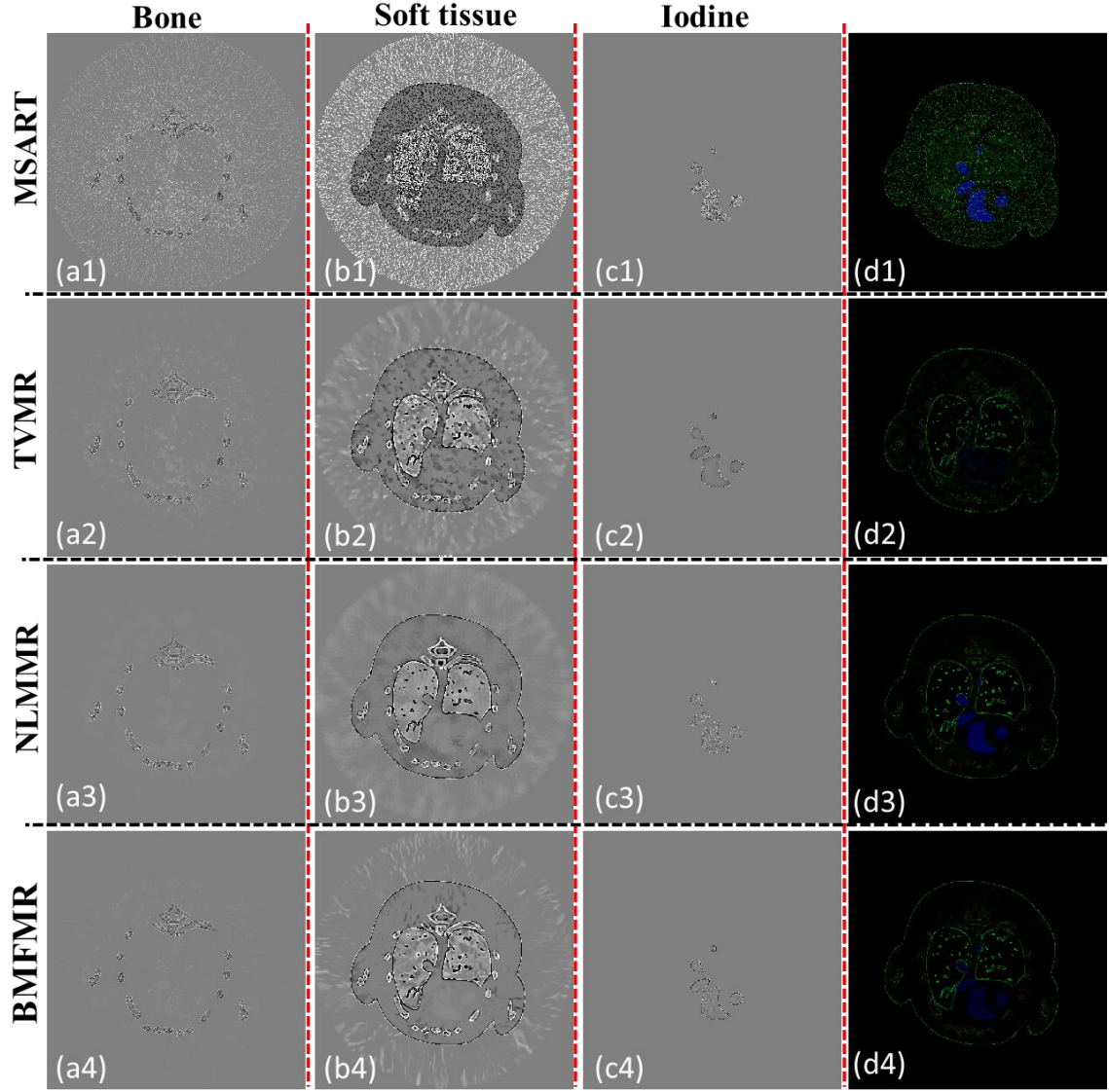

Figure 5. The difference images between the ground truths and reconstructed material images. The display windows of bone, water and iodine are [-0.2, 0.2], [-0.2, 0.2] and [-0.001, 0.001], respectively. The 4$^{th}$ column is the color rendering image of difference.

Table II. Quantitative evaluation results of different reconstruction methods.

| Material | Index | MSART | TVMR | NLMMR | BMFMR |
|---|---|---|---|---|---|
| | RMSE($10^{-2}$) | 1.43 | 1.37 | **1.32** | **1.32** |
| Bone  | SSIM         | 0.584 | 0.952 | 0.979 | **0.987** |
|       | PSNR(unit: db) | 31.50 | 37.29 | 37.58 | **37.61** |
| Water | RMSE($10^{-2}$) | 6.43  | 6.26  | 5.87  | **5.56** |
|       | SSIM         | 0.330 | 0.709 | 0.768 | **0.842** |
|       | PSNR(unit: db) | 16.43 | 24.06 | 24.63 | **24.99** |
| Iodine| RMSE($10^{-2}$) | 2.35  | 1.77  | 1.69  | **1.62** |
|       | SSIM         | 0.988 | **0.999** | 0.994 | 0.998 |
|       | PSNR(unit: db) | 70.43 | 75.03 | 75.42 | **75.80** |

*c). Convergence analysis*

To investigate the convergences of the proposed algorithms, the convergent curves in terms of RMSEs of water *vs.* iteration numbers from all the algorithms are given in **Fig. 6**. From **Fig. 6**, one can see that all the optimized algorithms can converge to a stable solution with the smallest RMSE. In fact, for the BMFMR methods, the RMSE decreases rapidly at first, and then it is subsequently stable after 30 iterations. More rigid theoretical analysis is provided in Appendix A.2.

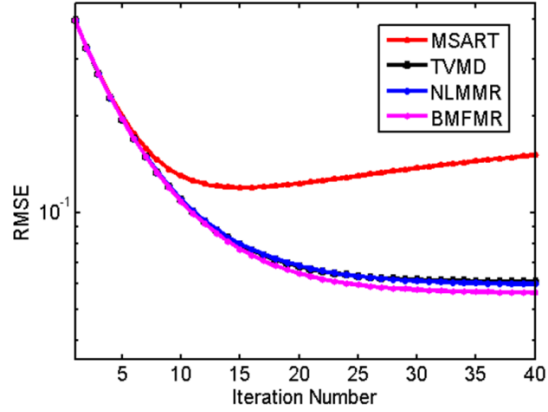

Figure 6. RMSEs curves of water vs. iteration numbers.

*d). Computational cost*

As far as the computational cost is concerned, the proposed algorithms mainly divide into three subroutines: the multi-energy projection dataset decomposition, back-projection reconstruction and regularization. The whole computational cost depends on the number of materials. The higher the number of the materials is, the larger the time consumption is. Regarding the regularization term BMF, the computational cost depends on a series of parameter settings such as the patch size, search neighborhood size, the patch sliding step, *etc*. In this study, all the algorithms are programmed by Matlab (version 2017b) on a PC (i7-6700, 32.0 GB memory). For this system configuration, it takes 2.19 seconds for the FBP algorithm to reconstruct image for each energy bin. Here, each iteration of the MSART, TVMR, NLMMR and BMFMR methods consume 101.46, 102.41, 480.73 and 114.96 seconds respectively. Obviously, the NLM regularization term requires more time than the TV and BMF regularizations.

*B. Physical Phantom Experiments*

A physical phantom containing three basis materials (i.e., water, iodine and aluminum) is scanned by an experimental spectral CT system in Capital Normal University (CNU). As shown in **Fig. 7**, the spectral CT system includes a micro-focus x-ray source (YXLON, 225kV) and a flat-panel PCD (Xcounter, XC-Hydra). This type of XCounter detector is the High Resolution (HR) configuration of SANTIS 0804. The sensor is manufactured of CdTe. Here, the PCD consists of 2048 detector cells and each of them covers 0.1 mm. To reduce noise, every 4 cells are combined to a form a low resolution detector with 512 cells. Because the PCD only has two energy channels, the projections containing 4 different energy bins with 360 views are obtained by scanning the phantom multiple times and the voltage settings of the x-ray source is 137 kVp. The distances starting from the x-ray source to object center and the PCD are 182.68 mm and 440.50 mm, respectively. Thus, the radius of field of view (FOV) is 41.3 mm. In this study, the size of each reconstructed material image is 512×512 and each pixel covers an area of 0.162×0.162 mm$^2$.

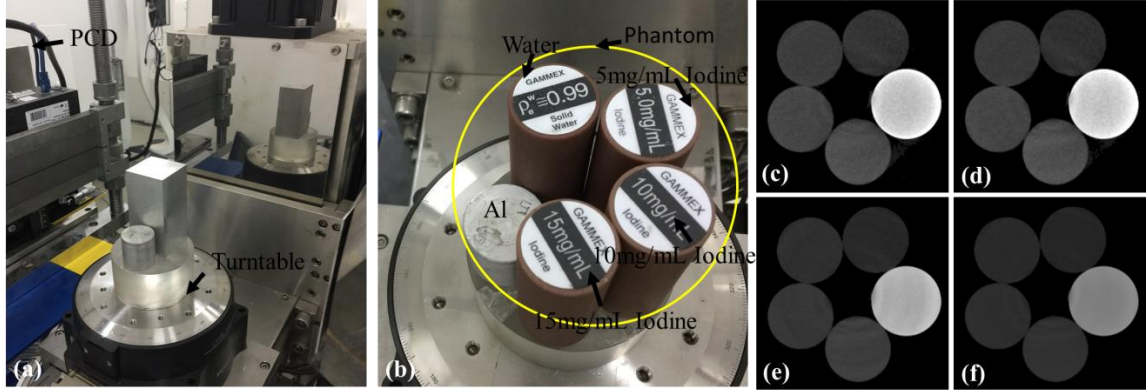

Figure 7. Setups of physical phantom experiments. (a) is the spectral CT system and (b) is the phantom containing five cylinders and each of them represents different basis material or different concentrations of iodine solution, (c)-(f) represent spectral CT images reconstructed by FBP from four different energy bins.

The x-ray energy spectrum plays an important role in our proposed methods and the error between the estimated and ground truth may compromise the material reconstruction results. In this study, the detected energy spectrum is estimated by manually adjusting the threshold with a small step and appropriate post-processing techniques (see **Fig. 8**). Because there is no imaging object, it can be used to estimate the emitted spectrum after the bow-tie filtering. From **Fig. 8**, the x-ray spectrum starting from 13 keV to 137 keV is divided into 4 energy channels, *i.e.*, [13, 25], (25, 33], (33, 48] and (48, 137], and the reconstructed spectral CT images by FBP are given in **Fig. 7 (c)-(f)**. Since the physical phantom contains the aluminum, there exists the x-ray beam hardening artifacts in the reconstructed images. The one-step based iteration material decomposition methods can reduce x-ray beam hardening artifacts with accurate x-ray emitting spectrum. Again, the accuracy of estimated spectrum can direct affect the accuracy of the final reconstructed results. The closer the estimated spectrum to the truth is, the higher the accuracy of reconstructed materials is. **Fig. 9** shows the reconstructed three basis materials. From **Fig. 9**, it can be seen that some of the aluminum is wrongly classified as water. Compared with the results of MSART, TVMR and NLMMR, the BMFMR can obtain the highest accuracy of material decomposition. To validate this conclusion, the ROIs 1, 2 and 3 are extracted in the aluminum, water and iodine contrast agent materials to compute the indexes of RMSE, PSNR and SSIM. Here, the ground truth of ROIs 1 and 2 are equal to 1.0 in theory. Correspondingly, the ground truth of ROI-3 is 15mg/mL. The quantitative evaluation results are listed in Table III. From **Table III**, the BMFMR can obtain the smallest RMSE and PSNR values in all ROIs. In general, the BMFMR can obtain higher SSIM values comparing with other competitors. Especially, the RMSE values of ROI 1 are large. This is because the energy spectrum used in this study is not accurate and it compromises the final material reconstruction results. Due to the inaccuracy of energy spectrum, 15mg/mL and 10mg/mL iodine concentrations can be observed in all methods, but 5mg/mL iodine concentration is difficult to see in all reconstructed results.

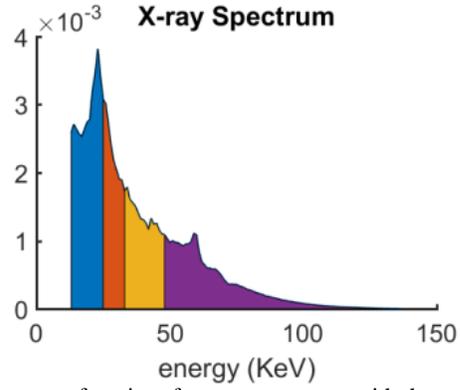

Figure 8. Estimated spectrum of a micro-focus x-ray source with the voltage setting 137kVp.

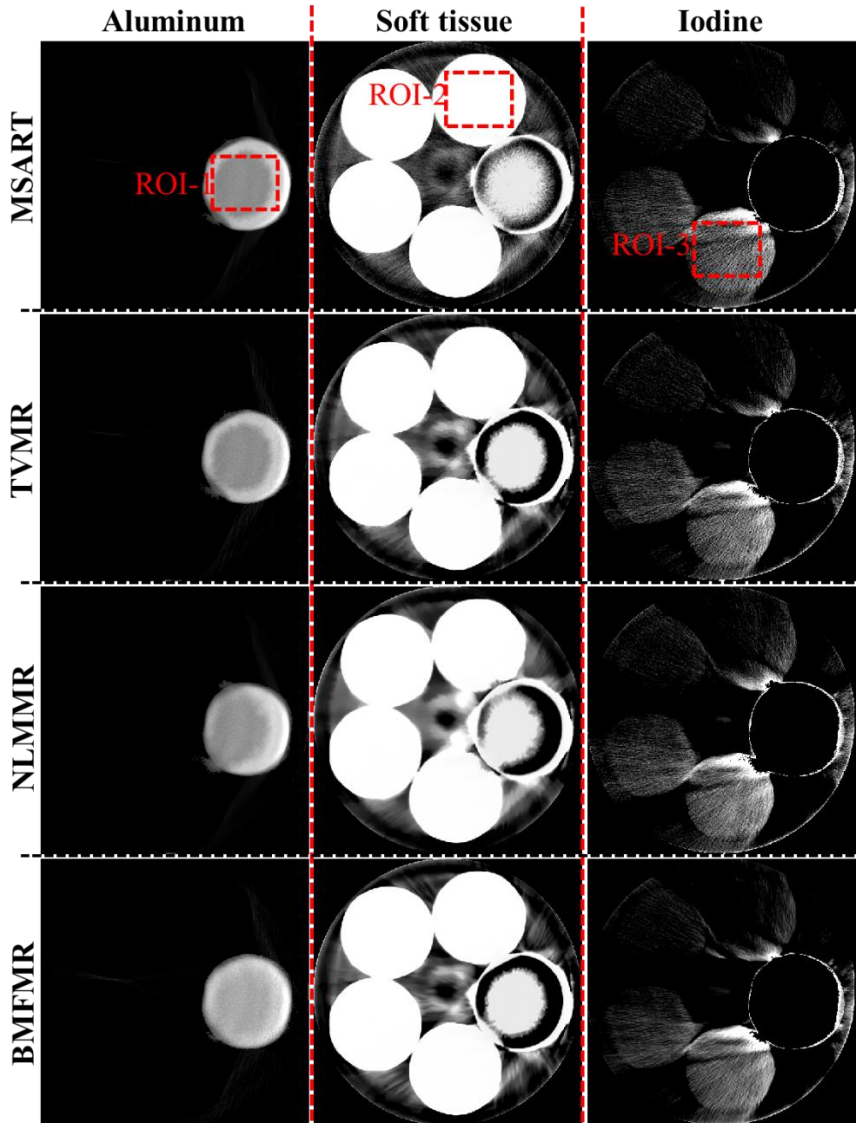

Figure 9. Physical phantom experiment results. The display windows of aluminum, water and iodine are [0 1], [0 1] and [0.00002 0.003], respectively.

| Material | Index | MSART | TVMR | NLMMR | BMFMR |
|---|---|---|---|---|---|
| ROI-1 | RMSE | 0.435 | 0.432 | 0.392 | **0.322** |
| | SSIM | 0.756 | 0.819 | 0.826 | **0.838** |
| | PSNR(unit: db) | 7.239 | 7.299 | 8.133 | **9.849** |
| ROI-2 | RMSE($10^{-3}$) | 4.979 | 4.319 | 3.523 | **3.332** |
| | SSIM ($10^{-1}$) | 9.742 | **9.946** | 9.943 | 9.941 |
| | PSNR(unit: db) | 46.06 | 47.29 | 49.06 | **49.55** |
| ROI-3 | RMSE($10^{-3}$) | 5.424 | 5.196 | 4.780 | **4.479** |
| | SSIM | 0.783 | 0.794 | 0.805 | **0.810** |
| | PSNR(unit: db) | 45.32 | 45.69 | 46.41 | **46.98** |

Table III. Quantitative evaluation results of ROIs 1-3.

## *V. Conclusions and discussions*

To reconstruct basis material images directly from the spectral CT projection datasets, we first propose an MSART algorithm. Compared with the E-ART method for DECT (Zhao *et al.*, 2015), the innovations of MSART are demonstrated in the following three ways. First, the MSART method is proposed by considering the spectral CT imaging model and Taylor's first-order expansion rather than the basic idea of ART. Second, the E-ART ignores the implicate relationship between projection decomposition and image reconstruction. The MSART method elaborates this relationship by introducing image constraint $\boldsymbol{P} = (\boldsymbol{A}\boldsymbol{\mathcal{F}})^T$ in Eq. (14). Third, the MSART model is strictly deduced by minimizing a higher-order Taylor expansion rather than by a simple analysis for E-ART, which can refer to Eqs. (8) - (12) and ***Theorem 1*** in the Appendix *A.1*. Next, we incorporate the TV, NLM and BMF regularizations into the MSART model to generate the TVMR, NLMMR and BMFMR algorithms. This can help to improve the robustness of the MSART method and preserve the material image edge information. Both numerical simulation and physical phantom experiment results demonstrate that the reconstructed material images by the BMFMR method have higher accuracy compared with those reconstructed by the MSART, TVMR and NLMMR methods.

In this study, we mainly focus on one-step based material decomposition method for spectral CT. To further compare the proposed BMFMR method with indirect method, the material decomposition results using indirect method are shown in **Fig. 10**. Here, FBP technique is employed to achieve spectral CT images and then the direct inversion method (Wu *et al.*, 2019; Zhang *et al.*, 2017) is used to obtain the final reconstruction results. Compared with the results in Fig. 3, it can be seen from Fig. 10 that many pixels of iodine contrast agent and soft tissue are wrongly classified as bone component. Besides, there are many outlier artifacts in the iodine contrast agent by comparing with those obtained from the BMFMR method. In terms of soft tissue, the clearer edges may be achieved by the direct inversion. However, the result still contains noises within the lung regions.

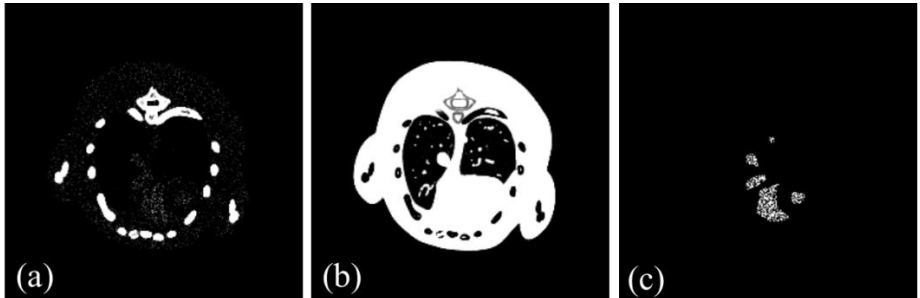

Figure 10. Reconstructed basis material images of simulation experiment. The display windows of bone, water and iodine are [0.012 0.1], [0.35 0.80] and [0.011 0.012], respectively.

Since the accuracy of the estimated energy spectrum can compromise the accuracy of the final material reconstructions, estimation of the energy spectrum is the largest limitation of the proposed methods in practical applications. There are many advanced techniques to estimate highly accurate x-ray energy spectrum, including material calibration (Wu *et al.*, 2012) and so on. In this study, we just computed the energy spectrum by manually adjusting the threshold and performing a difference operation. This may result in compromised accuracy of the reconstructed material images. It is necessary to use a more accurate spectrum to obtain higher accuracy of material components. However, because the x-ray energy spectrum estimation is a typical ill-posed problem, it is difficult to estimate accurate x-ray energy spectrum in practice. Besides, the overlap of different energy bins is common in spectral CT imaging (Schmidt *et al.*, 2017), which also is a challenge to estimate x-ray energy spectrum. There are lots of methods available to estimate x-ray energy spectrum, including truncated singular value decomposition (TSVD)(Armbruster *et al.*, 2004), expectation-maximization (EM) (Sidky et al., 2005), prior truncated singular value decomposition (PTSVD) (Leinweber *et al.*, 2017) and so on. However, the x-ray energy spectrum estimation is beyond the scope of this study, and it is scheduled in our next plan. Because material reconstruction methods consider each x-ray path independently, they can suppress beam-hardening artifacts in the reconstructed material images. However, the noise levels will be magnified by decomposing the average projection dataset into independent basis material contributions. In fact, the selection of material decomposition methods may depend on specific applications, such as the number of basis materials, scan speed and noise level from multi-spectral acquisition. Especially, the computational cost for performing iterative reconstruction should be considered to satisfy some clinical requirements. Again, if any non-iterative based method can provide similar results as the iterative methods, the non-iteration based method would be a good choice. Although exciting results have been achieved by incorporating the TV, NLM and BMF regularizations into the MSART model, there is still some room for improvement. For example, the quality of the reconstructed material images by the BMFMR method is still poor in the case of $10^5$ photons per x-ray path although it is much better than that obtained by the TVMR and NLMMR methods. In addition, although the piece-wise constant mouse phantom is sufficient to validate and evaluate the proposed algorithms, it is necessary to further design more complicated phantoms with non-ideal spectrum to demonstrate the practical values for our proposed algorithm in the future. To further improve the accuracy of material reconstruction, in the future we will consider more constraints to optimize the solution. In our study, the parameters are selected empirically based on extensive experiments. In our follow-up work, we will also try some automatic schemes for parameter optimization, including but not limited to the L-curve (Belge *et al.*, 2002), Morozov's discrepancy principle (Xie and Zou, 2002), semismooth Newton (Clason et al., 2010), generalized cross validation function (Jansen, 2015), adapted parameter selection strategy (Frikel, 2013) and so on. A comprehensive evaluation will also be performed to further optimize the parameters.

In conclusion, based on Taylor's first order expansion theory, we first propose an original one-step material reconstruction algorithm MSART for spectral CT. An optimized BMFMR method is further developed to well preserve image edges and improve the robustness of MSART. This will be extremely useful for basis material reconstruction for spectral CT application.

## *Acknowledgement*

The authors are grateful to Mr. Joshua Lojzim at UMass Lowell for his help on language editing.

## *Appendix*
### A.1. Minimization of Y($P$) defined by Eq. (17)
For the objective function Y($P$) defined by Eq. (17), we have the following Theorem.

***Theorem 1:*** Let $\lambda > 0$, $Y(\boldsymbol{P})$ is a strictly convex function with respect to $\boldsymbol{P}$ in a feasible set $\boldsymbol{P} \in \mathcal{P}$. Then, $Y(\boldsymbol{P})$ can reach its global minimizer by the following iterative formula

$$\left(\boldsymbol{P}_{\#1}^{(k+1)}, \cdots, \boldsymbol{P}_{\#L}^{(k+1)}\right) = \left(\boldsymbol{P}_{\#1}^{(k)} - \beta_1 \left(\boldsymbol{\Theta}_{\#1\#}^{(k)}{}^T \boldsymbol{\Theta}_{\#1\#}^{(k)} + \lambda \boldsymbol{\mathcal{I}}\right)^{-1} \boldsymbol{S}_1^{(k)}\left(\overline{\boldsymbol{Q}}_{\#1} - \boldsymbol{Q}_{\#1}^{(k)}\right), \cdots, \boldsymbol{P}_{\#L}^{(k)} - \beta_1 \left(\boldsymbol{\Theta}_{\#L\#}^{(k)}{}^T \boldsymbol{\Theta}_{\#L\#}^{(k)} + \lambda \boldsymbol{\mathcal{I}}\right)^{-1} \boldsymbol{S}_L^{(k)}\left(\overline{\boldsymbol{Q}}_{\#L} - \boldsymbol{Q}_{\#L}^{(k)}\right)\right),$$

where $\beta_1$ is a constant to control the step size and $k$ indicates the iteration number.

***Proof:*** $Y(\boldsymbol{P})$ can be rewritten as

$$Y(\boldsymbol{P}) = \sum_{\ell=1}^{L} \left\{ \left\| \boldsymbol{\Theta}_{\#\ell\#}^{(k)}\left(\boldsymbol{P}_{\#\ell} - \boldsymbol{P}_{\#\ell}^{(k)}\right) + \boldsymbol{S}_\ell^{(k)}\left(\overline{\boldsymbol{Q}}_{\#\ell} - \boldsymbol{Q}_{\#\ell}^{(k)}\right) \right\|_F^2 + \lambda \left\| ((\boldsymbol{A}_{\ell\#}\boldsymbol{\mathcal{F}})^T)^{(k)} - \boldsymbol{P}_{\#\ell} \right\|_F^2 \right\}. \quad (A.1)$$

Because the measured data from each x-ray path $\ell$ are independent, to obtain the optimized solution for Eq. (A.1), we can estimate an arbitrary $\ell^{th}$ $(1 \leq \ell \leq L)$ x-ray $\boldsymbol{P}_{\#\ell}$ independently. Without loss of generality, let us compute

$$\frac{\partial Y(\boldsymbol{P})}{\partial \boldsymbol{P}_{\#\ell}} = \boldsymbol{\Theta}_{\#\ell\#}^{(k)}{}^T \left(\boldsymbol{\Theta}_{\#\ell\#}^{(k)}\left(\boldsymbol{P}_{\#\ell} - \boldsymbol{P}_{\#\ell}^{(k)}\right) + \boldsymbol{S}_\ell^{(k)}\left(\overline{\boldsymbol{Q}}_{\#\ell} - \boldsymbol{Q}_{\#\ell}^{(k)}\right)\right) + \lambda\left(\boldsymbol{P}_{\#\ell} - ((\boldsymbol{A}_{\ell\#}\boldsymbol{\mathcal{F}})^T)^{(k)}\right). \quad (A.2)$$

Let $\partial Y(\boldsymbol{P})/\partial \boldsymbol{P}_{\#\ell} = 0, (1 \leq \ell \leq L)$, and we have

$$\boldsymbol{\Theta}_{\#\ell\#}^{(k)}{}^T \left(\boldsymbol{\Theta}_{\#\ell\#}^{(k)}\left(\boldsymbol{P}_{\#\ell} - \boldsymbol{P}_{\#\ell}^{(k)}\right) + \boldsymbol{S}_\ell^{(k)}\left(\overline{\boldsymbol{Q}}_{\#\ell} - \boldsymbol{Q}_{\#\ell}^{(k)}\right)\right) + \lambda\left(\boldsymbol{P}_{\#\ell} - ((\boldsymbol{A}_{\ell\#}\boldsymbol{\mathcal{F}})^T)^{(k)}\right) = 0. \quad (A.3)$$

It can be evolved into

$$\boldsymbol{\Theta}_{\#\ell\#}^{(k)}{}^T \boldsymbol{\Theta}_{\#\ell\#}^{(k)} \boldsymbol{P}_{\#\ell} + \lambda \boldsymbol{P}_{\#\ell} = \boldsymbol{\Theta}_{\#\ell\#}^{(k)}{}^T \boldsymbol{\Theta}_{\#\ell\#}^{(k)} \boldsymbol{P}_{\#\ell}^{(k)} - \boldsymbol{S}_\ell^{(k)}\left(\overline{\boldsymbol{Q}}_{\#\ell} - \boldsymbol{Q}_{\#\ell}^{(k)}\right) + \lambda((\boldsymbol{A}_{\ell\#}\boldsymbol{\mathcal{F}})^T)^{(k)}. \quad (A.4)$$

Note that $\boldsymbol{P}_{\ell\#}^{(k)} = ((\boldsymbol{A}_{\ell\#}\boldsymbol{\mathcal{F}})^T)^{(k)}$. Eq. (A.4) can be simplified as

$$\left(\boldsymbol{\Theta}_{\#\ell\#}^{(k)}{}^T \boldsymbol{\Theta}_{\#\ell\#}^{(k)} + \lambda \boldsymbol{\mathcal{I}}\right) \boldsymbol{P}_{\#\ell} = \left(\boldsymbol{\Theta}_{\#\ell\#}^{(k)}{}^T \boldsymbol{\Theta}_{\#\ell\#}^{(k)} + \lambda \boldsymbol{\mathcal{I}}\right) \boldsymbol{P}_{\#\ell}^{(k)} - \boldsymbol{S}_\ell^{(k)}\left(\overline{\boldsymbol{Q}}_{\#\ell} - \boldsymbol{Q}_{\#\ell}^{(k)}\right), \quad (A.5)$$

where $\boldsymbol{\mathcal{I}}$ represents the identity transform whose size is the same as $\boldsymbol{\Theta}_{\#\ell\#}^{(k)}{}^T \boldsymbol{\Theta}_{\#\ell\#}^{(k)}$. Since $\lambda > 0$, the matrix $\boldsymbol{\Theta}_{\#\ell\#}^{(k)}{}^T \boldsymbol{\Theta}_{\#\ell\#}^{(k)} + \lambda \boldsymbol{\mathcal{I}}$ is always reversible. The solution of $\boldsymbol{P}_{\#\ell}$ can be iteratively updated as follow

$$\boldsymbol{P}_{\#\ell}^{(k+1)} = \boldsymbol{P}_{\#\ell}^{(k)} - \beta_1 \left(\boldsymbol{\Theta}_{\#\ell\#}^{(k)}{}^T \boldsymbol{\Theta}_{\#\ell\#}^{(k)} + \lambda \boldsymbol{\mathcal{I}}\right)^{-1} \boldsymbol{S}_\ell^{(k)}\left(\overline{\boldsymbol{Q}}_{\#\ell} - \boldsymbol{Q}_{\#\ell}^{(k)}\right), \quad (A.6)$$

where $\left(\boldsymbol{\Theta}_{\#\ell\#}^{(k)}{}^T \boldsymbol{\Theta}_{\#\ell\#}^{(k)} + \lambda \boldsymbol{\mathcal{I}}\right)^{-1}$ is the inverse matrix of the $\left(\boldsymbol{\Theta}_{\#\ell\#}^{(k)}{}^T \boldsymbol{\Theta}_{\#\ell\#}^{(k)} + \lambda \boldsymbol{\mathcal{I}}\right)$ and $\beta_1$ is a relaxation factor in $(0,2)$. Thus, we can obtain such a current stationary point $\boldsymbol{\wp} = \left(\boldsymbol{P}_{1\#}^{(k+1)}, \cdots, \boldsymbol{P}_{L\#}^{(k+1)}\right)$, which can be given as

$$\boldsymbol{\wp} = \begin{pmatrix} \boldsymbol{P}_{\#1}^{(k)} - \beta_1 \left(\left(\boldsymbol{\Theta}_{\#1\#}^{(k)}{}^T \boldsymbol{\Theta}_{\#1\#}^{(k)} + \lambda \boldsymbol{\mathcal{I}}\right)\right)^{-1} \boldsymbol{S}_1^{(k)}\left(\overline{\boldsymbol{Q}}_{\#1} - \boldsymbol{Q}_{\#1}^{(k)}\right) \\ \vdots \\ \boldsymbol{P}_{\#L}^{(k)} - \beta_1 \left(\left(\boldsymbol{\Theta}_{\#L\#}^{(k)}{}^T \boldsymbol{\Theta}_{\#L\#}^{(k)} + \lambda \boldsymbol{\mathcal{I}}\right)\right)^{-1} \boldsymbol{S}_L^{(k)}\left(\overline{\boldsymbol{Q}}_{\#L} - \boldsymbol{Q}_{\#L}^{(k)}\right) \end{pmatrix}^T. \quad (A.7)$$

However, the convergence point of such a $\boldsymbol{\wp}$ is only a stationary point and it cannot be guaranteed as the global minimization point. To ensure the current point $\boldsymbol{\wp}$ is the optimization minimization point, it is necessary to compute the Hessian matrix $\boldsymbol{\mathcal{H}}$ of $Y(\boldsymbol{P})$

$$\mathcal{H} = \begin{bmatrix} \frac{\partial^2 Y(\boldsymbol{P})}{\partial \boldsymbol{P}_{\#1}\partial \boldsymbol{P}_{\#1}} & \cdots & \frac{\partial^2 Y(\boldsymbol{P})}{\partial \boldsymbol{P}_{\#1}\partial \boldsymbol{P}_{\#L}} \\ \vdots & \ddots & \vdots \\ \frac{\partial^2 Y(\boldsymbol{P})}{\partial \boldsymbol{P}_{\#L}\partial \boldsymbol{P}_{\#1}} & \cdots & \frac{\partial^2 Y(\boldsymbol{P})}{\partial \boldsymbol{P}_{\#L}\partial \boldsymbol{P}_{\#L}} \end{bmatrix}, \quad (A.8)$$

where $\mathcal{H} \in \mathcal{R}^{Z \times Z}$ and $Z = L \times N \times N$. Because each of x-ray is independent, that means

$$\frac{\partial^2 Y(\boldsymbol{P})}{\partial \boldsymbol{P}_{\#\ell}\partial \boldsymbol{P}_{\#j}} = \frac{\partial \left( \boldsymbol{\Theta}_{\#\ell\#}^{(k)}{}^T \left( \boldsymbol{\Theta}_{\#\ell\#}^{(k)}\left(\boldsymbol{P}_{\#\ell} - \boldsymbol{P}_{\#\ell}^{(k)}\right) + \left(\overline{\boldsymbol{Q}}_{\#\ell} - \boldsymbol{Q}_{\#\ell}^{(k)}\right)\right) + \lambda\left(\boldsymbol{P}_{\#\ell} - ((\boldsymbol{A}_{\ell\#}\boldsymbol{\mathcal{F}})^T)^{(k)}\right) \right)}{\partial \boldsymbol{P}_{\#j}} = \begin{cases} \boldsymbol{\Theta}_{\#\ell\#}^{(k)}{}^T \boldsymbol{\Theta}_{\#\ell\#}^{(k)} + \lambda \boldsymbol{\mathcal{I}}, & \ell = j \\ \boldsymbol{0}, & otherwise, \end{cases} \quad (A.9)$$

where $\boldsymbol{0}$ stands for a zero matrix in $\mathcal{R}^{N \times N}$. Thus, the Hessian matrix $\mathcal{H}$ can be further converted into

$$\mathcal{H} = \begin{bmatrix} \boldsymbol{\Theta}_{\#1\#}^{(k)}{}^T \boldsymbol{\Theta}_{\#1\#}^{(k)} + \lambda \boldsymbol{\mathcal{I}} & \cdots & \boldsymbol{0} \\ \vdots & \ddots & \vdots \\ \boldsymbol{0} & \cdots & \boldsymbol{\Theta}_{\#L\#}^{(k)}{}^T \boldsymbol{\Theta}_{\#L\#}^{(k)} + \lambda \boldsymbol{\mathcal{I}} \end{bmatrix} = \begin{bmatrix} \boldsymbol{\Theta}_{\#1\#}^{(k)}{}^T \boldsymbol{\Theta}_{\#1\#}^{(k)} & \cdots & \boldsymbol{0} \\ \vdots & \ddots & \vdots \\ \boldsymbol{0} & \cdots & \boldsymbol{\Theta}_{\#L\#}^{(k)}{}^T \boldsymbol{\Theta}_{\#L\#}^{(k)} \end{bmatrix} + \lambda \boldsymbol{\mathcal{I}}$$

$$= \begin{bmatrix} \boldsymbol{\Theta}_{\#1\#}^{(k)} & \cdots & \boldsymbol{0} \\ \vdots & \ddots & \vdots \\ \boldsymbol{0} & \cdots & \boldsymbol{\Theta}_{\#L\#}^{(k)} \end{bmatrix}^T \begin{bmatrix} \boldsymbol{\Theta}_{\#1\#}^{(k)} & \cdots & \boldsymbol{0} \\ \vdots & \ddots & \vdots \\ \boldsymbol{0} & \cdots & \boldsymbol{\Theta}_{\#L\#}^{(k)} \end{bmatrix} + \lambda \boldsymbol{\mathcal{I}}$$

$$= \mathcal{H}_1^T \mathcal{H}_1 + \lambda \boldsymbol{\mathcal{I}} > \boldsymbol{0}, \quad (A.10)$$

where $\lambda > 0$. Because the Hessian matrix $\mathcal{H}$ of $Y(\boldsymbol{P})$ is always greater than 0 and is a positive definite matrix, the global minimization point of $Y(\boldsymbol{P})$ exists and it can be iteratively determined by Eq. (A.6), which completes the proof of *Theorem 1*. ∎

## A. 2. Theoretical analysis of the BMFMR algorithm

Because the objective function is monotonically minimized in all the steps of the alternating optimization process, the value of the objective function should be monotonically decreased for the proposed BMFMR algorithm. Since Eq. (18) monotonically minimizes the objective function (17), according to *Theorem 1*, we have

$$\sum_{\ell=1}^{L} \left\{ \left\| \boldsymbol{\Theta}_{\#\ell\#}^{(k)}\left(\boldsymbol{P}_{\#\ell}^{(k+1)} - \boldsymbol{P}_{\#\ell}^{(k)}\right) + \boldsymbol{S}_\ell^{(k)}\left(\overline{\boldsymbol{Q}}_{\#\ell} - \boldsymbol{Q}_{\#\ell}^{(k)}\right) \right\|_F^2 + \lambda \left\| ((\boldsymbol{A}_{\ell\#}\boldsymbol{\mathcal{F}})^T)^{(k)} - \boldsymbol{P}_{\#\ell}^{(k+1)} \right\|_F^2 \right\}$$

$$\leq \sum_{\ell=1}^{L} \left\{ \left\| \boldsymbol{\Theta}_{\#\ell\#}^{(k-1)}\left(\boldsymbol{P}_{\#\ell}^{(k)} - \boldsymbol{P}_{\#\ell}^{(k-1)}\right) + \boldsymbol{S}_\ell^{(k-1)}\left(\overline{\boldsymbol{Q}}_{\#\ell} - \boldsymbol{Q}_{\#\ell}^{(k-1)}\right) \right\|_F^2 + \lambda \left\| ((\boldsymbol{A}_{\ell\#}\boldsymbol{\mathcal{F}})^T)^{(k-1)} - \boldsymbol{P}_{\#\ell}^{(k)} \right\|_F^2 \right\},$$

$$\forall k = 2,3 \ldots K. \quad (A.11)$$

Considering Eq. (29) monotonically minimizes the objective function (28a), according to the steepest descent theory, we can obtain

$$\left\| \boldsymbol{A}\boldsymbol{f}_n^{(k+1)} - \left(\boldsymbol{P}_{n\#}^{(k+1)}\right)^T \right\|_F^2 + \gamma_n \left\| \boldsymbol{f}_n^{(k+1)} - \boldsymbol{g}_n^{(k)} - \boldsymbol{t}_n^{(k)} \right\|_F^2$$

$$\leq \left\| \boldsymbol{A}\boldsymbol{f}_n^{(k)} - \left(\boldsymbol{P}_{n\#}^{(k+1)}\right)^T \right\|_F^2 + \gamma_n \left\| \boldsymbol{f}_n^{(k)} - \boldsymbol{g}_n^{(k)} - \boldsymbol{t}_n^{(k)} \right\|_F^2,$$

$$\forall n = 1, \ldots N. \quad (A.12)$$

As for Eq. (28b), it can be divided into a data fidelity term and a regularization term. On one hand, the $\boldsymbol{\Phi}_n(\boldsymbol{g}_n)$ can be considered as the 3D transform coefficient $\boldsymbol{\mathcal{W}}_n$. Thus, Eq. (28b) can be read as

$$\min_{\mathbf{g}_n} \left\| \mathbf{f}_n^{(k+1)} - \mathbf{g}_n - \mathbf{t}_n^{(k)} \right\|_F^2 \quad , s.t. \mathcal{W}_n \leq \mathbf{L}_n ,$$
$$\forall n = 1, \ldots N, \tag{A.13}$$

where $\mathbf{L}_n$ represents the 3D transform level. Clearly, the solution of problem (A.13) satisfies

$$\left\| \mathbf{f}_n^{(k+1)} - \mathbf{g}_n^{(k+1)} - \mathbf{t}_n^{(k)} \right\|_F^2 \leq \left\| \mathbf{f}_n^{(k+1)} - \mathbf{g}_n^{(k)} - \mathbf{t}_n^{(k)} \right\|_F^2 . \tag{A.14}$$

On the other hand, the hard thresholding method was employed to minimize problem (28b) in the regularization step. Thus, we have

$$\gamma_n \left\| \mathbf{f}_n^{(k+1)} - \mathbf{g}_n^{(k+1)} - \mathbf{t}_n^{(k)} \right\|_F^2 + \tau_n \left\| \Phi_n \left( \mathbf{g}_n^{(k+1)} \right) \right\|_0$$
$$\leq \gamma_n \left\| \mathbf{f}_n^{(k+1)} - \mathbf{g}_n^{(k)} - \mathbf{t}_n^{(k)} \right\|_F^2 + \tau_n \left\| \Phi_n \left( \mathbf{g}_n^{(k)} \right) \right\|_0 ,$$
$$\forall n = 1, \ldots N. \tag{A.15}$$

Noting that Eq. (30) monotonically reduces the value of (28c), we have

$$\left\| \mathbf{f}_n^{(k+1)} - \mathbf{g}_n^{(k+1)} - \mathbf{t}_n^{(k+1)} \right\|_F^2 \leq \left\| \mathbf{f}_n^{(k+1)} - \mathbf{g}_n^{(k+1)} - \mathbf{t}_n^{(k)} \right\|_F^2 , \forall n = 1, \ldots N. \tag{A.16}$$

From formulae (A.12)-(A.16), we can obtain

$$\left\| A \mathbf{f}_n^{(k+1)} - \left( \mathbf{P}_{n\#}^{(k+1)} \right)^T \right\|_F^2 + \gamma_n \left\| \mathbf{f}_n^{(k+1)} - \mathbf{g}_n^{(k+1)} - \mathbf{t}_n^{(k+1)} \right\|_F^2 + \tau_n \left\| \Phi_n \left( \mathbf{g}_n^{(k+1)} \right) \right\|_0$$
$$\leq \left\| A \mathbf{f}_n^{(k)} - \left( \mathbf{P}_{n\#}^{(k+1)} \right)^T \right\|_F^2 + \gamma_n \left\| \mathbf{f}_n^{(k)} - \mathbf{g}_n^{(k)} - \mathbf{t}_n^{(k)} \right\|_F^2 + \tau_n \left\| \Phi_n \left( \mathbf{g}_n^{(k)} \right) \right\|_0 . \tag{A.17}$$

Combining Eqs. (A.11) and (A.17), finally it shows that the BMFMR monotonically reduces its objective function. It is necessary to emphasize that the aforementioned monotonic reduction of the objective function cannot guarantee the convergence of the reconstruction algorithms and the BMFMR. Because the reconstruction model of Eq. (24) contains the $L_0$-norm minimization of 3D transform coefficients, the proof of the convergence of BMFMR algorithm is difficult, and thus out of the scope of this study.